\documentclass[10pt,journal,compsoc]{IEEEtran}
\usepackage[latin9]{inputenc}
\usepackage{url}
\usepackage{amstext}
\usepackage{graphicx}

\makeatletter

\providecommand{\tabularnewline}{\\}
\newcommand{\lyxdot}{.}

 \let\oldforeign@language\foreign@language
 \DeclareRobustCommand{\foreign@language}[1]{%
   \lowercase{\oldforeign@language{#1}}}

\ifCLASSOPTIONcompsoc
  \usepackage[nocompress]{cite}
\else
  \usepackage{cite}
\fi

\ifCLASSOPTIONcompsoc
  \usepackage[caption=false,font=footnotesize,labelfont=sf,textfont=sf]{subfig}
\else
  \usepackage[caption=false,font=footnotesize]{subfig}
\fi

\usepackage{amsmath}
\usepackage{amssymb}
\usepackage{colortbl}
\usepackage{array}
\usepackage{times}
\usepackage{ragged2e}

\def\etal{\emph{et~al}.~}
\def\figword{Fig.~}

\newcommand\colorsquare[2][black]{\textcolor{#1}{\rule{#2}{#2}}}

\makeatother

\begin{document}
\IEEEtitleabstractindextext{\justify{
\begin{abstract}
Humans implicitly rely on properties of the materials that make up
ordinary objects to guide our interactions. Grasping smooth materials,
for example, requires more care than rough ones, and softness is an
ideal property for fabric used in bedding. Even when these properties
are not purely visual (softness is a physical property of the material),
we may still infer the softness of a fabric by looking at it. We refer
to these visually-recognizable material properties as visual material
attributes. Recognizing visual material attributes in images can contribute
valuable information for general scene understanding and for recognition
of materials themselves. Unlike well-known object and scene attributes,
visual material attributes are local properties. ``Fuzziness'',
for example, does not have a particular shape. We show that given
a set of images annotated with known material attributes, we may accurately
recognize the attributes from purely local information (small image
patches). Obtaining such annotations in a consistent fashion at scale,
however, is challenging. We introduce a method that allows us to solve
this problem by probing the human visual perception of materials to
automatically discover unnamed attributes that serve the same purpose.
By asking simple yes/no questions comparing pairs of image patches,
we obtain sufficient weak supervision to build a set of attributes
(and associated classifiers) that, while being unnamed, serve the
same function as the named attributes, such as ``fuzzy'' or ``rough'',
with which we describe materials. Doing so allows us to recognize
visual material attributes without resorting to exhaustive manual
annotation of a fixed set of named attributes. Furthermore, we show
that our automatic attribute discovery method may be integrated in
the end-to-end learning of a material classification CNN framework
to simultaneously recognize materials and discover their visual material
attributes. Our experimental results show that visual material attributes,
whether named or automatically discovered, provide a useful intermediate
representation for known material categories themselves as well as
a basis for transfer learning when recognizing previously-unseen categories.
\end{abstract}

}

}

\title{Recognizing Material Properties from Images}

\author{Gabriel Schwartz and Ko Nishino, \IEEEmembership{Senior Member, IEEE}\thanks{The authors are with the Department of Computer Science, Drexel University,
Philadelphia, PA, 19104.\protect \\
E-mail: \protect\url{{gbs25,kon}@drexel.edu}}}


\maketitle
\IEEEdisplaynontitleabstractindextext

\IEEEpeerreviewmaketitle

\section{Introduction}

\IEEEPARstart{P}{roperties} of the materials that appear in everyday
scenes inform many of the decisions we make when interacting with
things made from these materials. When cleaning a glass cup, for example,
we know not to drop it or it will break. Glass is also often smooth,
and we grasp it accordingly so it will not slip. Examples of material
properties include visual properties, such as glossiness or translucency,
as well as physical or tactile properties, such as hardness or roughness.
We can see the presence of these material properties simply by looking
at ordinary images, suggesting that even non-visual properties can
be inferred from the visual appearance of the material. We refer to
such visually-recognizable material properties as visual material
attributes. Recognizing these visual material attributes in images
would allow us to better understand and interact with the scenes in
which materials appear.

\begin{figure}
\begin{centering}
\includegraphics[width=0.95\columnwidth]{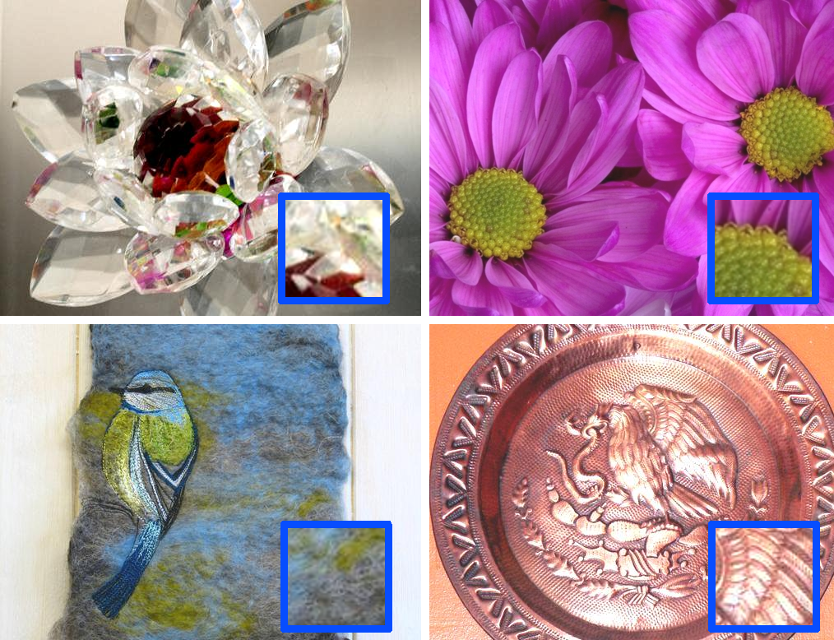}
\par\end{centering}
\caption{\label{fig:Local-material-properties}Materials are unique in that
they have many characteristic locally-recognizable visual attributes.
Wool, for example, often appears ``fuzzy'', and metals are typically
``shiny''. In this work we show that material attributes can be
recognized from purely local information, that we can discover these
visual attributes automatically (again from purely local information),
and that such a process can be applied to large-scale material image
datasets via an end-to-end formulation in a Convolutional Neural Network.}
\end{figure}

Recognizing visual material attributes in images, however, is particularly
challenging. Unlike much prior work in object, face, and scene attribute
recognition~\cite{Farhadi2009,Ferrari2008,Kumar2008,Patterson2012},
material attributes are local properties, as can be seen in Figure~\ref{fig:Local-material-properties}.
While scene, object, and face attributes are typically associated
with a characteristic shape and fixed spatial extent, material attributes
are not. A car that is ``sporty'' will have a typically sleek and
aerodynamic shape, and if a face has ``large eyes'', such an attribute
is defined by the spatial extent of the eyes relative to the face.
If a carpet is ``fuzzy'', however, the fuzziness is not associated
with any characteristic shape or scale, nor is it necessarily a consequence
of the object involved (some carpets are not fuzzy).

In a preliminary experiment, we show that given a fully-supervised
dataset containing material annotations, a set of named semantic material
attributes (such as ``fuzzy'', ``organic'', or ``transparent''),
and sparse per-pixel labels for these attributes, we can indeed recognize
visual material properties from small local image patches. Such a
method can be applied in a sliding window fashion to produce per-pixel
predictions for material attributes in arbitrary images. The accuracy
of these predictions supports our claim that material attributes are
indeed locally-recognizable and can be recognized at the per-pixel
level. Furthermore, we show that these attributes alone can be used
to recognize materials in a way that separates the material itself
from the surrounding context such as objects and scenes.

The primary drawback of a straightforward fully-supervised approach,
such as the one described above, is that it requires a set of consistent
annotations for the material attributes. Some material attributes
are intuitive and challenging to precisely define. If we wish to scale
the annotation process to multiple annotators, we can no longer assume
that a single person is providing attribute labels in a consistent
and complete fashion. Additionally, such a method implicitly depends
on the choice of a fixed set of named material attributes. This restriction
gives us no way of evaluating if the chosen set of material attributes
is complete or if there are possibly more attributes that we may implicitly
associate with materials.

Rather than assuming that we can exhaustively describe the set of
attributes humans associate with materials, we show that we can instead
directly probe the human visual perception of materials using simple
yes/no questions. Using the answers to these questions as a form of
weak supervision, we derive a method for discovering a set of unnamed
locally-recognizable visual material attributes that faithfully encodes
our own human perceptual representation of materials. Our method requires
only material annotations, and discovers unnamed attributes with the
same desirable properties as the fixed named material attributes we
described previously, while using only a small amount of easily-collected
weak supervision.

Our attribute discovery method requires only simple supervision and
eliminates the need to manually define a set of named material attributes
for full supervision. The training process is, however, still not
ideal for application to modern large-scale image databases. Working
well with small amounts of training data is a benefit, but we would
ideally like to leverage recent advances in large-scale end-to-end
learning as well. To this end, we show that the same material attributes
can in fact be discovered within a Convolutional Neural Network (CNN)
framework focused on material recognition in local image patches (the
Material Attribute/Category CNN, MAC-CNN). This enables us to take
advantage of potentially larger material datasets. We also find interesting
parallels with the material representation in the human material recognition
process as observed in neuroscience~\cite{Hiramatsu2011,Goda2014}.
In contrast to the intermediate representations formed by previous
attribute methods, the human material recognition process (and our
MAC-CNN) produces a perceptual representation, i.e.~visual material
attributes, as a side-product of material category recognition. Our
results show that we are able to discover similar perceptual attributes
using the MAC-CNN, and we additionally demonstrate the usefulness
of perceptual material attributes for transfer learning.

\section{Related Work}

In this paper we discuss the recognition of material properties from
images. There is much recent work in the areas of attribute recognition,
material recognition, and human visual perception that is relevant
to our work; we discuss these findings below.

\subsection{Attributes}

\subsubsection{Fully-Supervised Attributes}

Fully-supervised visual attributes have been widely used in object
and scene recognition, but largely at the image or scene level. Ferrari
and Zisserman~\cite{Ferrari2008} introduced a generative model for
certain pattern and color attributes, such as ``dots'', or ``stripes''.
The attributes described in their model focus on texture and color,
but are not material attributes. A paper cup, for example, may have
stripes painted on it, but ``striped'' is not a property of the
paper itself; it is in fact a property of the cup. Kumar~\etal\cite{Kumar2008}
proposed a face search engine with their attribute-based FaceTracer
framework. FaceTracer uses SVM and AdaBoost to recognize attributes
within fixed facial regions. Such fixed regions are not present in
materials, which may take on an arbitrary shape unlike the objects
which they make up. Farhadi~\etal\cite{Farhadi2009} applied attributes
to the problem of object recognition. Their results showed an improvement
in accuracy over a basic approach using texture features. Lampert~\etal\cite{Lampert2009}
also showed that attributes transfer information between disjoint
sets of classes. These results suggest that attributes can serve as
an intermediate representation for recognition of the categories which
exhibit them. Patterson and Hays~\cite{Patterson2012} showed that
they could recognize a variety of visual attributes, some of which
were in fact general material categories. Their work, however, was
not an explicit attempt at recognizing materials.

With a few exceptions, the majority of past attribute recognition
methods produced single image-wide predictions. Given an image of
a zebra, for example, the attribute prediction would be ``striped''.
As our goal is to recognize materials within local regions, we cannot
rely on such global attribute predictions.

\subsubsection{Weakly-Supervised Attributes}

The attributes described above were all fully-supervised or ``semantic''
attributes. A semantic attribute is one to which we can assign a name
like ``round'' or ``transparent''. While these attributes were
shown to be useful, it is difficult to quantify the completeness and
consistency of any given attribute set: does the set of attributes
contain everything that could help recognize the target categories,
and can the appearance (for visual attributes) be agreed upon by a
variety of annotators? Semantic attributes are also task-specific
and must be manually defined for each new recognition task.

To address the issues inherent to semantic attributes, a number of
unsupervised or weakly-supervised attribute discovery methods have
been proposed. Berg~\etal\cite{Berg2010} described a framework
for automatically learning object attributes from web data (images
and associated text). This approach learns some localized attributes.
The required text annotations are, however, image-wide and do not
guarantee locality. Patterson and Hays~\cite{Patterson2012} also
proposed a process to discover and recognize scene-wide attributes
in natural images. While they are able to discover a large amount
of attributes, their learned attributes are not local. Rastegari~\etal\cite{Rastegari2012}
learn a binary attribute representation (binary codes) for images.
As with most existing methods, however, these attributes are image-wide
and not local. Cimpoi~\etal\cite{Cimpoi2013} demonstrated a method
for learning an arbitrary set of describable texture attributes based
on terms derived from psychological studies. As noted by Adelson~\cite{Adelson2001},
texture is only one component of material appearance, and cannot alone
describe our perception of materials. Though their results demonstrate
impressive performance on the FMD, their learned attributes apply
only globally. Most relevant to our work are the attribute discovery
methods of Akata~\etal\cite{Akata2013} and Yu~\etal\cite{Yu2013}.
Akata~\etal\cite{Akata2013} formulated attribute discovery as a
label embedding problem. Yu~\etal\cite{Yu2013} proposed a two-step
procedure for discovering and classifying attributes based on a similarity
matrix. They computed a distance matrix using Euclidean distances
in the raw feature space of labeled image patches. In contrast, we
embed the material categories in an attribute space derived from our
own human visual perception of material similarity.

\subsection{Material Recognition}

Adelson~\cite{Adelson2001} first suggested materials as a distinct
concept from objects or simple textures when discussing ``things
vs.~stuff''. ``Things'' refers to objects, which have been the
focus of much prior work under the field of object recognition. Adelson
points out that the world does not just consist of discrete objects,
but also includes ``stuff'', substances without a natural shape
or fixed spatial extent. Ice cream is one example of ``stuff'' that
is not an object but is still a recognizable concept in images. While
materials are not equivalent to the ``stuff'' discussed in his work,
the work does lay the foundation for material recognition as a vision
problem.

The first collection of material category images for classification
originated in Sharan~\etal\cite{Sharan2009} where they introduced
a new image database (the \textbf{F}lickr \textbf{M}aterials \textbf{D}atabase
or \textbf{FMD}) containing images from the photo sharing website
Flickr. The FMD contains a set of images each with a single material
annotation and corresponding mask identifying the presence of that
material. Building on the FMD, Liu~\etal\cite{Liu2010} created
a framework to recognize these material categories using a modified
LDA probabilistic topic model. Hu~\etal\cite{Hu2011} improved upon
the state-of-the-art FMD accuracy using kernel descriptors and large-margin
nearest neighbor distance metric learning. Their experiments showed
that providing explicit object detection information to material category
recognition results in a large improvement in accuracy. Sharan~\etal\cite{Sharan2013}
later showed that without information associated with objects (such
as the object shape), performance degrades significantly (from 57.1\%
to 42.6\%). Specifically, they note that their material category recognition
method depends heavily on non-local features such as edge contours.
Given that materials exhibit distinct locally-recognizable visual
attributes (as we show), it follows that we should be able to recognize
them in a way that does not suffer from reduced accuracy in the absence
of context.

All prior work discussed above produces a single category prediction
for each input image. This inherently assumes that there is only one
material of interest in the image, a very restrictive assumption.
To relax this assumption, recent work (including some of our own preliminary
work~\cite{Schwartz2017}) focuses on dense prediction: providing
a material category for each pixel in the input image. Bell~\etal
introduced the OpenSurfaces~\cite{Bell2013} and MINC~\cite{Bell2015}
datasets to aid in the training of dense material recognition models.
With MINC they also describe a simple modification of the VGG CNN
architecture of Simonyan and Zisserman~\cite{Simonyan2015} to predict
their material categories at each pixel.

\subsection{Material Perception and Convolutional Neural Networks}

As the final step in the scaling of material attribute learning, we
discover perceptual material attributes within Convolutional Neural
Networks (CNNs)~\cite{LeCun1990}. Convolutional neural networks
are general non-linear models that apply a set of convolution kernels
to an image in an hierarchical fashion to generate a category probability
vector. The kernel weights are model parameters that are set via non-linear
optimization (generally Stochastic Gradient Descent) to attempt to
maximize the likelihood of a set of training data.

Recently, Shankar~\etal\cite{Shankar2015} proposed a modified CNN
training procedure to improve attribute recognition. Their ``deep
carving'' algorithm provides the CNN with attribute pseudo-label
targets, updated periodically during training. This causes the resulting
network to be better-suited for attribute prediction. Escorcia~\etal\cite{Escorcia2015}
show that known semantic attributes can also be extracted from a CNN.
They show that attributes depend on features in all layers of the
CNN, which will be particularly relevant to our investigation of perceptual
material attributes in CNNs. ConceptLearner, proposed by Zhou~\etal\cite{Zhou2015}
uses weak supervision, in the form of images with associated text
content, to discover semantic attributes. These attributes correspond
to terms within the text that appear in the images. All of these frameworks
predict a single set of attributes for an entire image, as opposed
to the per-pixel attributes which we introduce.

At the intersection of neuroscience and computer vision, Yamins~\etal\cite{Yamins2014}
find that feature responses from high-performing CNNs can accurately
model the neural response of the human visual system in the inferior
temporal (IT) cortex (an area of the human brain that responds to
complex visual stimuli). They perform a linear regression from CNN
feature outputs to IT neural response measurements and find that the
CNN features are good predictors of neural responses despite the fact
that the CNN was not explicitly trained to match the neural responses.
Their work focuses on object recognition CNNs, not materials. Hiramatsu~\etal\cite{Hiramatsu2011}
take functional magnetic resonance imaging (fMRI) measurements and
investigate their correlation with both direct visual information
and perceptual material properties (similar to the material traits
of~\cite{Schwartz2013}) at various areas of the human visual system.
They find that pairwise material dissimilarities derived from fMRI
data correlate best with direct visual information (analogous to pixels)
at the lower-order areas and with perceptual attributes at higher-order
areas. Goda~\etal\cite{Goda2014} obtain similar findings in non-human
primates. These studies suggest the existence of perceptual attributes
in human material recognition, but do not actually derive a process
to extract them from novel images.

\section{\label{sec:Visual-Material-Traits}Visual Material Traits}

\begin{figure*}
\begin{centering}
\includegraphics[height=0.11\paperheight]{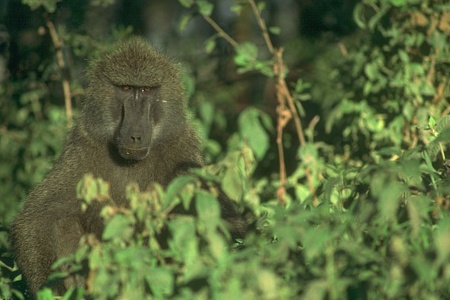} \includegraphics[height=0.11\paperheight]{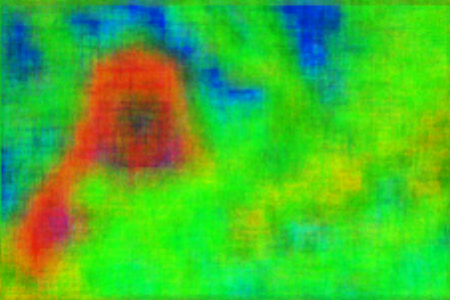}
\includegraphics[height=0.11\paperheight]{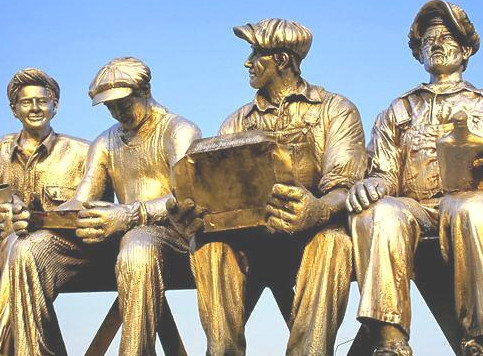}
\includegraphics[height=0.11\paperheight]{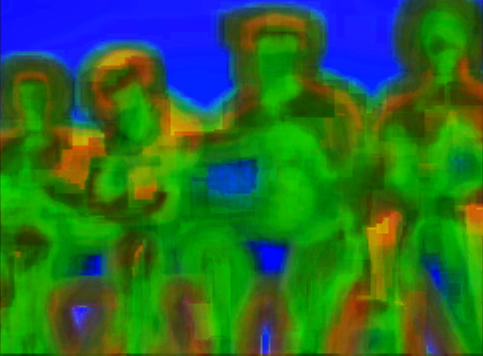}
\par\end{centering}
\begin{centering}
\hspace{20pt}%
\begin{tabular}{ccc}
\includegraphics[width=0.03\columnwidth]{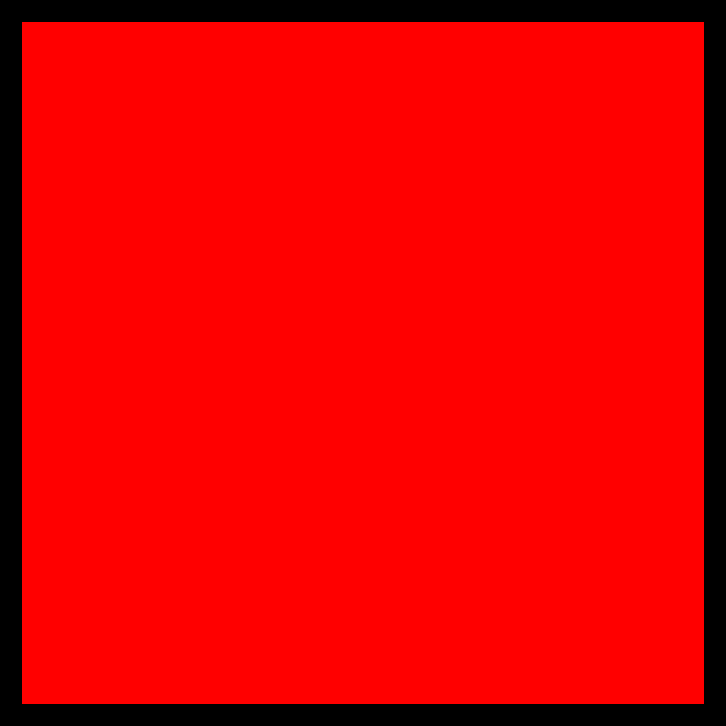}\hspace{3pt}Fuzzy & \includegraphics[width=0.03\columnwidth]{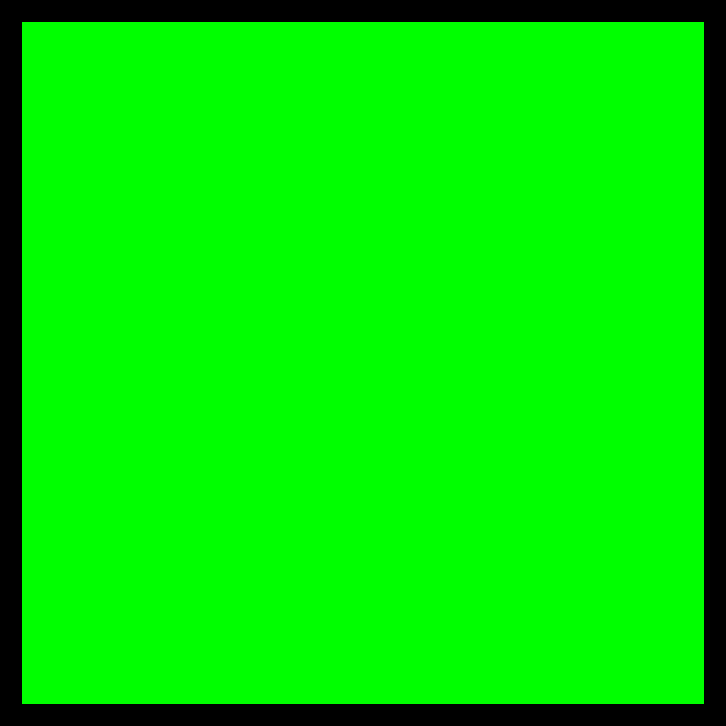}\hspace{3pt}Organic & \includegraphics[width=0.03\columnwidth]{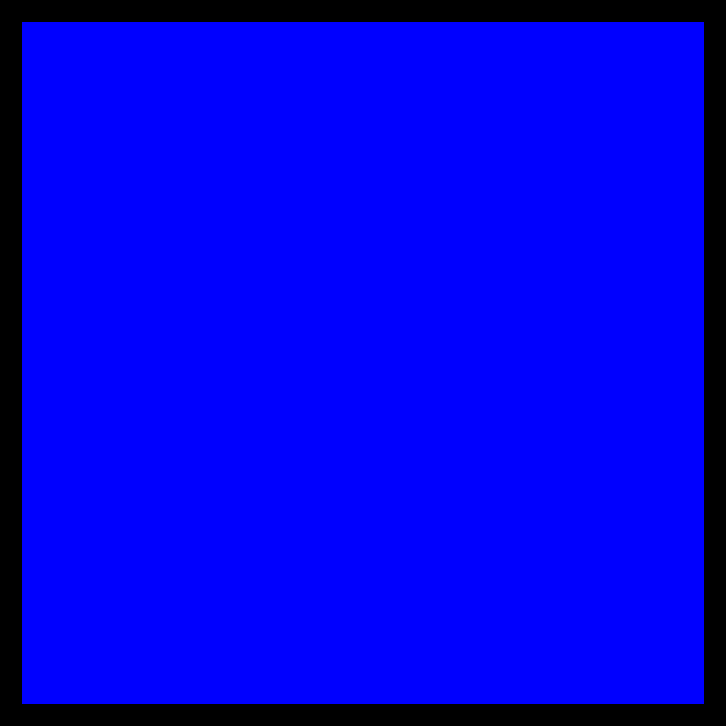}\hspace{3pt}Smooth\tabularnewline
\end{tabular}\hspace{90pt}%
\begin{tabular}{ccc}
\includegraphics[width=0.03\columnwidth]{images/red_box}\hspace{3pt}Shiny & \includegraphics[width=0.03\columnwidth]{images/green_box}\hspace{3pt}Metallic & \includegraphics[width=0.03\columnwidth]{images/blue_box}\hspace{3pt}Smooth\tabularnewline
\end{tabular}
\par\end{centering}
\caption{\label{fig:recognized-traits}Per-pixel material trait predictions
on unseen test images. Each RGB color channel corresponds to the predicted
probability of a single selected material trait. The recognized traits
clearly divide the images into regions of similar material appearance.
Of particular importance is the fact that we can successfully recognize
material properties like ``fuzzy'' on the left, despite the fact
that our training data did not include any animals. The shape of each
statue on the right is that of a person, but we see from the recognized
material traits that the material is in fact metal.}
\end{figure*}

As preliminary experiment to show that materials do in fact exhibit
locally-recognizable visual properties, we use a set of named visual
material attributes (visual material traits~\cite{Schwartz2013}).
By manually annotating images from the FMD with per-pixel masks for
13 material traits, we can then train a set of Randomized Decision
Forest~\cite{Breiman2001} classifiers to predict material traits
in small local regions. Experimental results show that visual material
traits can be recognized very accurately from small ($32\times32$)
image patches, some traits as high as 93.1\%, with an average accuracy
of 78.4\%. \figword\ref{fig:recognized-traits} shows sample per-pixel
recognition results for a selection of material traits.

\begin{figure}
\begin{centering}
\includegraphics[width=0.31\columnwidth]{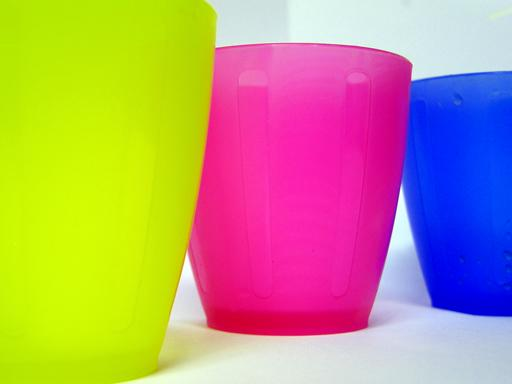}\includegraphics[width=0.31\columnwidth]{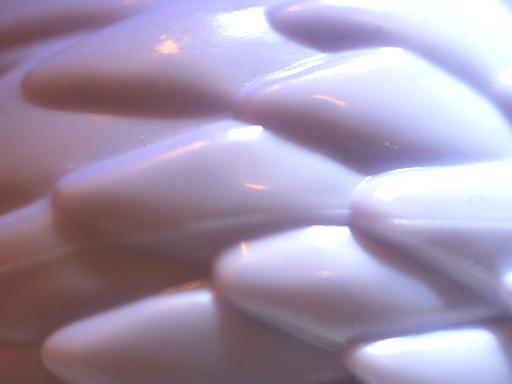}
\includegraphics[width=0.31\columnwidth]{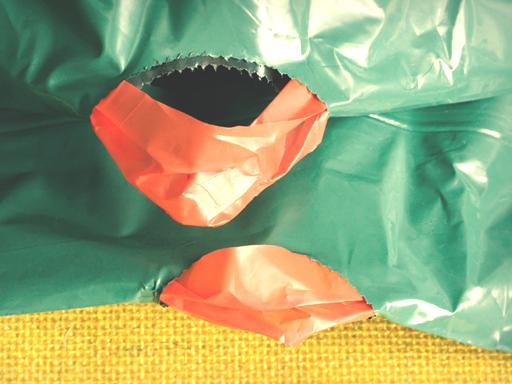}
\par\end{centering}
\vspace{2pt}
\begin{centering}
\includegraphics[width=0.31\columnwidth]{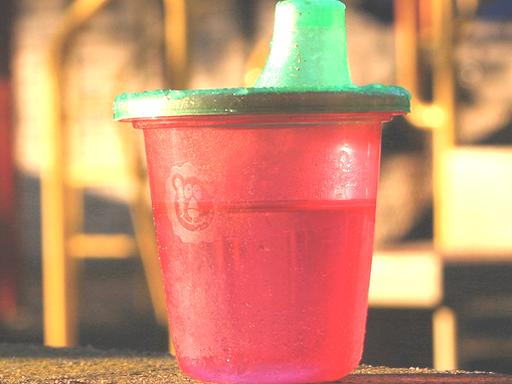} \includegraphics[width=0.31\columnwidth]{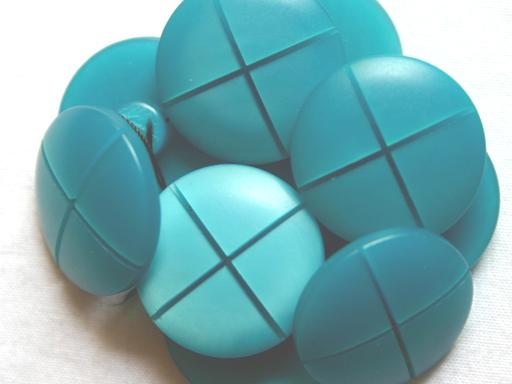}
\includegraphics[width=0.31\columnwidth]{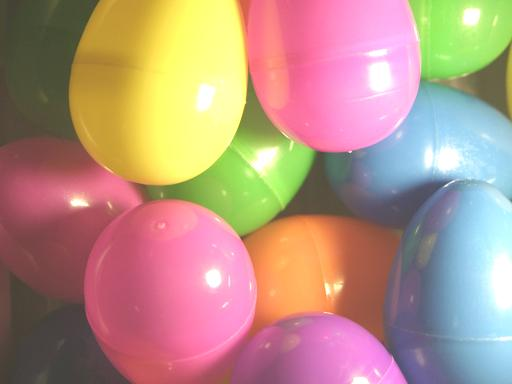}
\par\end{centering}
\caption{\label{fig:Intra-class-variability}Materials such as the plastic
in these images exhibit a wide range of appearances depending on the
object and scene. Despite this, we can intuitively recognize visual
attributes (smooth and translucent, for example) shared across different
instances of the material. In our preliminary work we show that these
attributes, which we refer to as visual material traits, are locally-recognizable
and can be used to recognize material categories like plastic or metal.}
\end{figure}

Materials, for example fabric, plastic, or metal, can be challenging
to recognize due to the large variation in appearance between instances
of the same material. Despite this, looking at the images in \figword\ref{fig:Intra-class-variability},
one can see that plastic tends to have visual material attributes
that are associated with a distinct appearance, such as ``smooth''
and ``translucent''. Our key observation is that these visual material
attributes are recognizable even when the surrounding objects and
scenes are not visible. We expect that as a result, we should be able
to recognize materials themselves from visual material traits given
only small local image patches.

Viewing the full set of material traits as an intermediate representation,
we may aggregate them within regions to describe materials. To exploit
the material properties found in locally-recognized material traits,
we treat the distribution of material traits in a region as an image
descriptor and generate a per-image material category prediction.
We do so by extracting a number of patches within each material region
and using the distributions of traits across these patches as features
for a histogram intersection SVM~\cite{Barla2003}. The average accuracy
of this method on 50-50 splits of FMD images is 49.2\%. While this
is not higher than the accuracy of methods such as~\cite{Sharan2013}
that implicitly use context, it is significantly higher than both
human and algorithmic performance in the absence of context (38.7-46.9\%
and 33.8-42.6\% respectively respectively~\cite{Sharan2013}). Furthermore,
material traits learned from one dataset can be recognized and used
to extract material information from an entirely different set, showing
that the representation generalizes well.

\section{\label{sec:Material-Attribute-Discovery}Material Attribute Discovery}

\begin{figure}
\begin{centering}
\includegraphics[width=0.12\columnwidth]{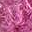} \includegraphics[width=0.12\columnwidth]{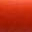}
\includegraphics[width=0.12\columnwidth]{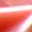} \includegraphics[width=0.12\columnwidth]{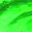}
\includegraphics[width=0.12\columnwidth]{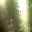}\includegraphics[width=0.12\columnwidth]{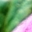}
\includegraphics[width=0.12\columnwidth]{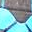}
\par\end{centering}
\vspace{2pt}
\begin{centering}
\includegraphics[width=0.12\columnwidth]{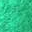} \includegraphics[width=0.12\columnwidth]{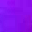}
\includegraphics[width=0.12\columnwidth]{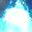} \includegraphics[width=0.12\columnwidth]{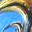}
\includegraphics[width=0.12\columnwidth]{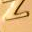} \includegraphics[width=0.12\columnwidth]{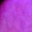}
\includegraphics[width=0.12\columnwidth]{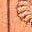}
\par\end{centering}
\vspace{2pt}
\begin{centering}
\includegraphics[width=0.12\columnwidth]{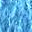} \includegraphics[width=0.12\columnwidth]{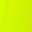}
\includegraphics[width=0.12\columnwidth]{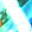} \includegraphics[width=0.12\columnwidth]{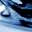}
\includegraphics[width=0.12\columnwidth]{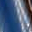} \includegraphics[width=0.12\columnwidth]{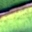}
\includegraphics[width=0.12\columnwidth]{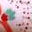}
\par\end{centering}
\caption{\label{fig:Local-trait-patches}Sample material image patches. Asking
annotators to merely ``describe'' the patches is an ambiguous question.
Patches may look similar even though the annotator does not have a
concrete word to define the similarity. We instead ask only for binary
visual similarity decisions.}
\end{figure}

Material trait recognition relies on a set of fully labeled material
trait examples. This assumption hinders scaling the method to larger
training datasets. We also do not have a complete, mutually-agreeable
vocabulary for describing materials and their visual characteristics;
named material traits are merely one attempt at describing material
appearance. This makes scaling with multiple annotators difficult.
Considering the images in the first column of \figword\ref{fig:Local-trait-patches},
for instance, one annotator may call them fuzzy and others may call
them fluffy. People may also be inconsistent in annotating material
traits. Some may only annotate the patches in the second column as
smooth and others may only see them as translucent. Cimpoi~\etal\cite{Cimpoi2013}
alleviate these problems for texture recognition by preparing a pre-defined
vocabulary. They may do so by focusing on apparent texture patterns
like stripes and dots. Materials underlie these texture patterns (\emph{i.e.},
the stripes or dots on a plastic cup are still plastic) and do not
follow such a vocabulary.

Our goal is to discover a set of attributes that exhibit the desirable
properties of material traits. We want to achieve this without relying
on fully-supervised learning. Known material traits, such as ``smooth''
or ``rough,'' represent visual properties shared between similar
materials. We expect that attributes that preserve this similarity
will satisfy our goal. We propose to define a set of attributes based
on the perceived distances between material categories. By working
with distances rather than similarities, we avoid any need to assume
a particular similarity function. For this, we obtain a measurement
of these distances from human annotations.

From a high-level perspective, our attribute discovery consists of
three steps:
\begin{enumerate}
\item Measure perceptual distances between materials
\item Define an attribute space based on perceptual distances
\item Train classifiers to reproduce this space from image patches
\end{enumerate}
Defining perceptual distance between material categories poses a challenge.
If each material had a single typical appearance (\emph{e.g.,} if
metal was always shiny and gray), we could simply compute the difference
between these typical appearances. This is not the case. Materials
may exhibit a wide variety of appearances, even sharing appearances
between categories (what we refer to as material appearance variability).
An image patch from a leaf, for example, may appear similar to certain
fabrics or plastics.

\subsection{\label{subsec:Measuring-Perceptual-Distances}Measuring Perceptual
Distances}

Directly measuring distances via human annotation would be ideal,
as we have an intuitive understanding of the differences between materials.
As Sharan~\etal\cite{Sharan2013} showed, this understanding persists
even in the absence of object cues. It is, however, also a difficult
task to obtain these distance. Given two query image patches, annotators
would have to decide how different the patches look on a consistent
quantitative scale. We would instead like to ask simple questions
that can be reliably answered.

We propose that instead of asking how different patches look, we reduce
the question to a binary one: ``Do these patches look different or
not?'' We assume that this will give us sufficient information to
obtain consistent and sensible perceptual information. Our underlying
assumption for this claim is that if a pair of image patches look
similar, they do so as a result of at least one shared visual material
trait.

To transform a set of binary similarity annotations into pairwise
distances, we represent each material as a point defined by the average
probabilities of similarity to each material category. The pairwise
distances between these points define the material perceptual distance
matrix. This process treats each material category as a point in a
space of typical (but not necessarily realizable) material appearances.
The resulting distance between a pair of materials depends on joint
similarity with all material categories, including the pair in question,
and is thus robust to material appearance variability.

Formally, given a set of $N$ reference images with material category
$c_{n}\in\left\{ 1\ldots K\right\} $, we obtain binary similarity
decisions $\mathbf{s}_{n}\in\left\{ 0,1\right\} ^{K}$ for each reference
image against a set of sample images from each category. We represent
each material category in the space of typical material category appearances
as $K$-dimensional vectors $\mathbf{p}_{k}$:\vspace{-5pt}
\begin{equation}
\mathbf{p}_{k}=\frac{1}{N_{k}}\sum_{n|c_{n}=k}\mathbf{s}_{n}\text{,}\label{eq:mat_cat_location}
\end{equation}
where $N_{k}=\left|\left\{ c_{n}|c_{n}=k\right\} \right|$. Entries
$d_{kk^{\prime}}$ in the $K\times K$ pairwise distance matrix $D$
are then defined as:

\vspace{-10pt}
\begin{equation}
d_{kk^{\prime}}=\left\Vert \mathbf{p}_{k}-\mathbf{p}_{k^{\prime}}\right\Vert _{2}\text{.}\label{eq:dist_mat_def}
\end{equation}

We obtain the required set of binary similarity annotations through
Amazon Mechanical Turk (AMT). Each task presents annotators with a
reference image patch of a given material category (unknown to the
annotator) and a row of random image patches, one from each material
category. We use patches from images of the 10 material categories
from the Flickr Materials Database of Sharan~\etal\cite{Sharan2009}.
Annotators are directed to select image patches that look similar
to the reference. Examples of suggested similar image patches are
given based on known material traits. Each set of patches is shown
to 10 annotators, and final results are obtained from a vote where
at least 5 annotators must agree that the patches look similar. We
collect similarity decisions for 10,000 reference image patches.

\begin{figure}
\begin{centering}
\includegraphics[width=0.99\columnwidth]{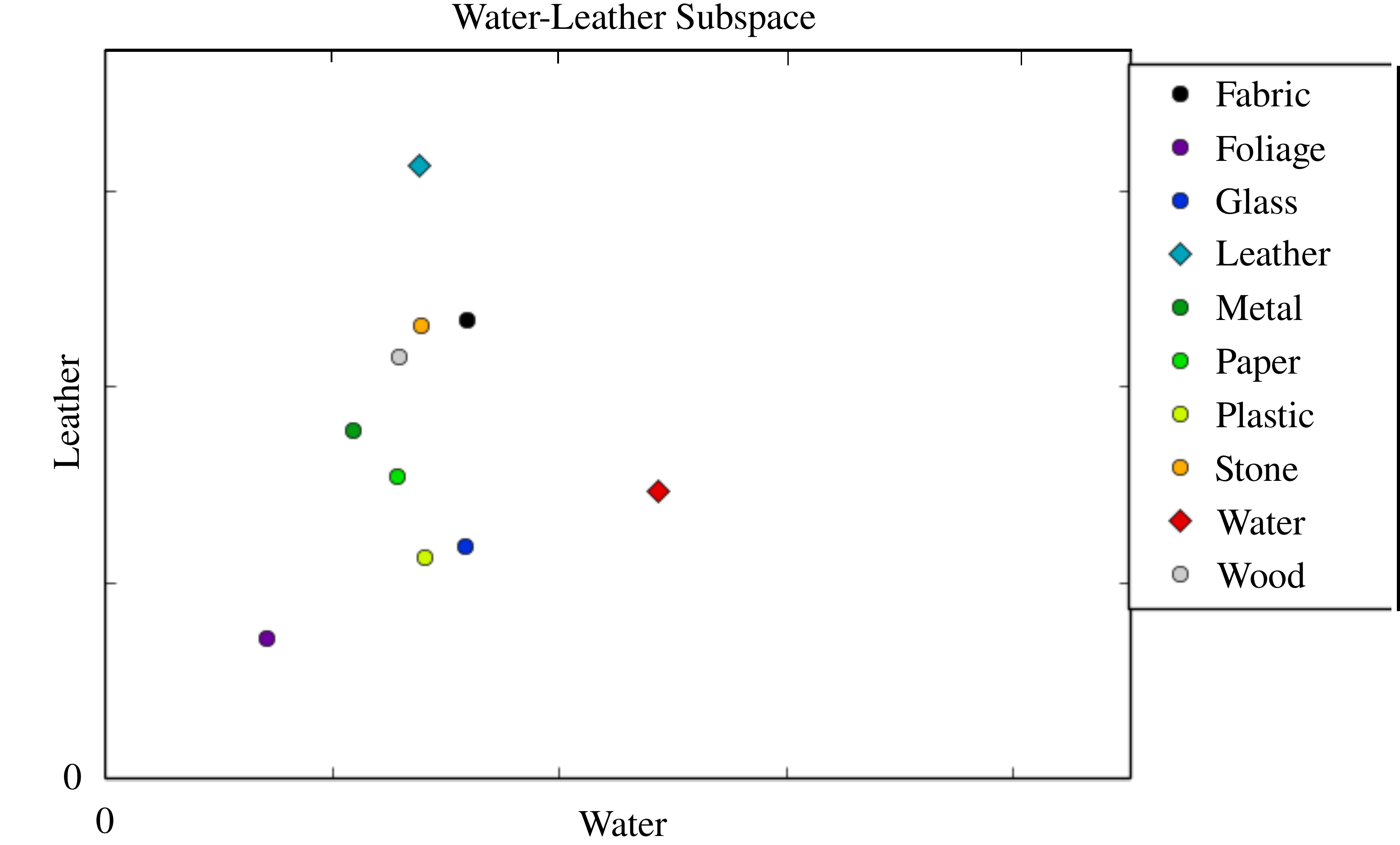}
\par\end{centering}
\caption{\label{fig:Example-projection-similarity-space}Projection of materials
into a 2D similarity subspace allow us evaluate our collected annotations.
Coordinates indicate perceived similarity to water and leather respectively.
The locations of the two material categories corresponding to the
axes are marked. We would expect that, in this case, water would lie
furthest along the ``water'' axis and likewise with leather. We
can also see that water occasionally looks like leather. Materials
with common visual properties, such as the smoothness of plastic and
glass, tend to lie close to each other. Finally, no material appears
strongly similar to both leather and water. This is expected as the
two materials do generally exhibit different appearances.}
\end{figure}

The 2D projection in \figword\ref{fig:Example-projection-similarity-space}
shows that the similarity values obtained from the AMT annotations
agree with our own intuitive understanding of material appearance.
The plot shows the locations of material categories projected into
one 2D subspace of the 10-dimensional space of material appearance
similarity (each dimension corresponding to similarity with that material).
We would expect that the two materials corresponding to the axes of
their subspace will lie close to their respective axes. In this case,
water is most similar with itself, but is also similar to glass. Leather
is likewise most similar with itself, but also similar to fabric.

To show that we do in fact obtain a consistent distance matrix, we
compute the difference between the distance matrix computed with all
annotations versus that from only $n$ of the $N$ total annotations.
The difference drops quickly (within the first few hundred samples
of 10,000), showing that annotators agree on a single common set of
perceptual distances.

\subsection{\label{subsec:Defining-the-Material}Defining the Material Attribute
Space}

Discovering attributes given only a desired distance matrix poses
a challenge. First, we have no definitions for the attributes since
we are trying to discover them from examples of materials. Even if
we had such definitions, we have no knowledge of any association between
the attributes and the materials. A straightforward approach might
be to directly train classifiers to output attribute values that encode
the distance matrix. This would be a particularly under-constrained
problem given the aforementioned lack of labels or associations.

We instead propose to separate attribute association and classifier
learning into two steps. First, we discover attributes in an abstract
form by discovering a mapping between categories and attribute probabilities.
This mapping places each material category into an attribute space.
We ensure that the mapping preserves the pairwise perceptual material
distances, and then train classifiers to predict the presence of these
attributes on image patches.

As described in Section~\ref{subsec:Measuring-Perceptual-Distances},
we obtain a distance matrix $\mathbf{D}$ from crowdsourced similarity
answers for $K$ material categories $C=\left\{ 1\ldots K\right\} $.
Using $\mathbf{D}$, we find a mapping that indicates which attributes
are associated with which categories. The number of attributes we
discover is arbitrary, and we refer to it as $M$. The mapping is
encoded in the $K\times M$ category-attribute matrix $\mathbf{A}$.
We restrict values in $\mathbf{A}$ to lie in the interval $\left[0,1\right]$
so that we may treat them as conditional probabilities.

We impose two constraints on the category attribute mapping. $\mathbf{A}$
should map categories to attributes in a way that preserves the measured
distances in $\mathbf{D}$, and the mapping should contain realizable
values. If the values in $\mathbf{A}$ are not plausible, we will
not be able to recognize the attributes on image patches. For example,
one potential attribute mapping would be to assign each attribute
to a single category. Attribute recognition would then become material
category recognition on single image patches, which is not feasible.

We formulate the attribute discovery process as a minimization problem
over category-attribute matrices $\mathbf{A}$:

\vspace{-8pt}
\begin{equation}
\mathbf{A}^{*}=\arg\min_{\mathbf{A}}d\left(\mathbf{D};\mathbf{A}\right)+w_{A}\kappa_{A}\left(\mathbf{A}\right)\label{eq:A_learn_obj}
\end{equation}
with hyperparameter $w_{A}$. $d$ describes how well the current
estimate of $\mathbf{A}$ encodes the pairwise perceptual differences
between material categories, and $\kappa_{A}$ is a constraint that
makes the discovered attribute associations exhibit a realizable distribution.

The category-attribute matrix that best encodes the desired pairwise
distances will minimize the following term defined over rows $\mathbf{a}_{k}$
of the matrix $\mathbf{A}$:

\begin{equation}
d\left(\mathbf{D};\mathbf{A}\right)=\sum_{k,k^{\prime}\in C}\left(\left\Vert \mathbf{a}_{k}-\mathbf{a}_{k^{\prime}}\right\Vert _{2}-\mathbf{D}_{kk\prime}\right)^{2}\text{.}\label{eq:A_learn_pdist1}
\end{equation}

To discover realizable attributes, we encode our own prior knowledge
that recognizable attributes exhibit a particular distribution and
sparsity pattern. We observe that semantic attributes, specifically
visual material traits, have a Beta-distributed association with material
categories. Generally, a material category will either strongly exhibit
a trait or it will not exhibit it at all. Intermediate cases occur
when a material category exhibits a particularly wide variation in
appearance. Fabric, for example, sometimes has a clear ``woven''
pattern but, in the case of silk or other smooth fabrics, does not.
We would like the values in $\mathbf{A}$ to be Beta-distributed to
match the distribution of known material trait associations.

The canonical method for matching two distributions is to minimize
a divergence measure between them. To incorporate this into a minimization
formulation, we need a differentiable measurement for the unknown
empirical distribution of values in $\mathbf{A}$. We choose the KL-divergence
and Gaussian kernel density estimator. The Gaussian kernel density
estimate at point $p$ is:

\begin{equation}
q\left(p;\mathbf{A}\right)=\frac{1}{KM}\sum_{k,m}\left(2\pi h^{2}\right)^{-\frac{1}{2}}\exp\left\{ -\frac{\left(a_{km}-p\right)^{2}}{2h^{2}}\right\} \label{eq:KDE}
\end{equation}
The KL-divergence between the distribution of the values in the category-attribute
matrix $\mathbf{A}$ and the target Beta distribution $\beta\left(p;a,b\right)$
with $a=b=0.5$ can then be written as:
\begin{equation}
\kappa_{A}\left(\mathbf{A}\right)=\sum_{p\in P}\beta\left(p;a,b\right)\ln\left(\frac{\beta\left(p;a,b\right)}{q\left(p;\mathbf{A}\right)}\right)\text{.}\label{eq:KL_KDE}
\end{equation}

\subsection{\label{subsec:Training-a-Material}Training a Material Attribute
Classifier}

We now must derive classifiers that recognize the attributes defined
by the category-attribute mapping. As attributes are not defined semantically,
we cannot ask for further annotation to label training patches with
attributes. Instead, we propose a model and a set of constraints that
will enable us to predict our discovered attributes on material image
patches.

We do not know \emph{a priori }any particular semantics or structure
associated with the attributes, thus we model our attributes using
a general two-layer non-linear model~\cite{Cybenko1989}. We constrain
the predictions such that they reproduce the desired values in the
attribute matrix (in expectation) while also separating material categories
when possible.

Formally, given a training set of $N$ image patches represented by
$D$-dimensional raw feature vectors $\mathbf{x}_{n}$ with corresponding
material categories $c_{n}\in C$, we train a model $f$ with parameters
$\mathbf{\Theta}$ that maps an image patch to $M$ attribute probabilities:
$f\left(\mathbf{x}_{n};\mathbf{\Theta}\right):\mathbb{R}^{D}\rightarrow\left[0,1\right]^{M}$.
Given an intermediate layer with dimensionality $H$ and parameters
$\mathbf{W}_{1}\in\mathbb{R}^{H\times D}$, $\mathbf{W}_{2}\in\mathbb{R}^{M\times H}$,
$\mathbf{b}_{1}\in\mathbb{R}^{H}$, $\mathbf{b}_{2}\in\mathbb{R}^{M}$
the prediction for an instance $\mathbf{x}_{n}$ is defined as:
\begin{eqnarray}
f\left(\mathbf{x}_{n};\mathbf{\Theta}\right) & = & h\left(\mathbf{W}_{2}h\left(\mathbf{W}_{1}\mathbf{x}_{n}+\mathbf{b}_{1}\right)+\mathbf{b}_{2}\right)\nonumber \\
h\left(x\right) & = & \min\left(\max\left(x,0\right),1\right)\text{.}\label{eq:2-layer-model}
\end{eqnarray}
As additional regularization, used only during training, we mask out
a random fraction of the weights used in the model to discourage overfitting
(akin to dropout~\cite{Hinton2012}).

We formulate the full classifier training process as a minimization
problem:

\vspace{-15pt}
\begin{eqnarray}
\mathbf{\Theta}^{\star}=\arg\min_{\mathbf{\Theta}}r\left(\mathbf{X};\mathbf{A},\mathbf{\Theta}\right)+w_{1}\kappa\left(\mathbf{X};\mathbf{\Theta}\right)-\nonumber \\
w_{2}\pi\left(\mathbf{X};\mathbf{A},\mathbf{\Theta}\right)\text{,}\label{eq:classifier_minimization}
\end{eqnarray}
with hyperparameters $w_{1}$ and $w_{2}$. $r$ (Equation~\ref{eq:unary_term})
is a data term indicating the difference between predicted and expected
attribute probabilities. $\kappa$ and $\pi$ (Equations~\ref{eq:kl_kde_classifier}
and \ref{eq:pairwise_term}) are, respectively, constraints on the
the distribution of attribute predictions and on the pairwise separation
of material categories.

The category-attribute matrix encodes the probabilities that each
category will exhibit each attribute. We represent this in our classifier
training by matching the mean predicted probability for each attribute
to the given entry in the category-attribute matrix:

\vspace{-15pt}
\begin{equation}
r\left(\mathbf{X};\mathbf{A},\mathbf{\Theta}\right)=\sum_{k\in C}\left\Vert \mathbf{a}_{k}-\frac{1}{N_{k}}\sum_{i|c_{i}=k}f\left(\mathbf{x}_{i};\mathbf{\Theta}\right)\right\Vert _{2}^{2}\text{.}\label{eq:unary_term}
\end{equation}
Equation~\ref{eq:unary_term} directly encodes the desired behavior
of the classifier, but it alone is under-constrained. Each prediction
for each instance may take on any value so long as their mean matches
the target value.

\begin{figure*}
\begin{centering}
\hfill{}\includegraphics[width=0.8\columnwidth]{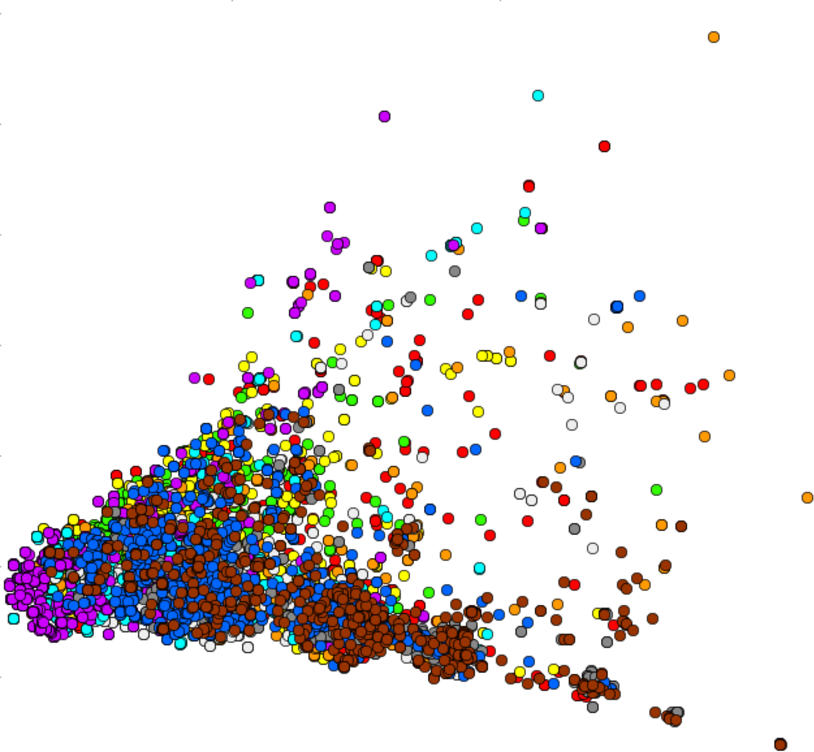}\hfill{}\includegraphics[width=0.8\columnwidth]{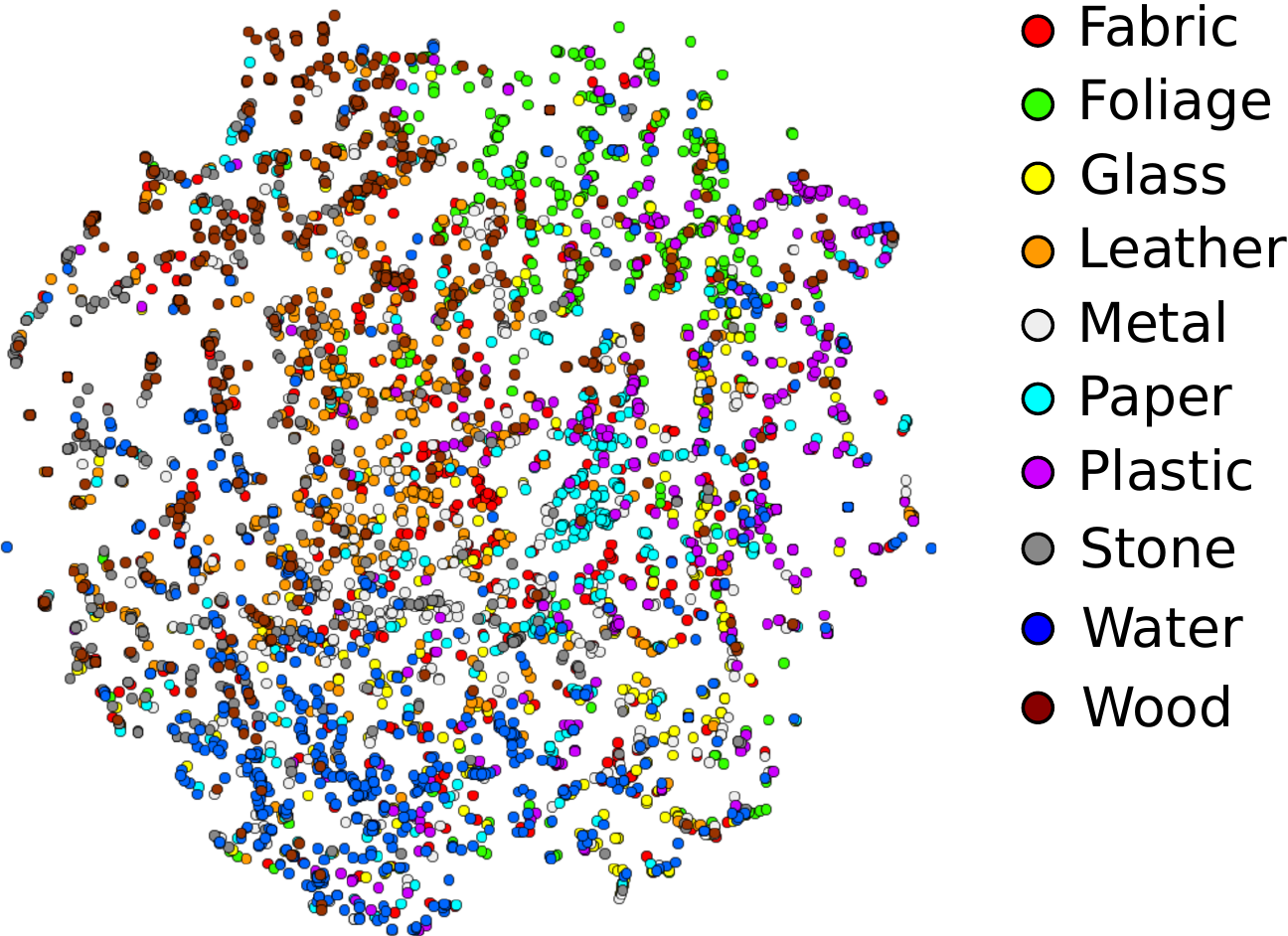}\hfill{}
\par\end{centering}
\hfill{}(a) Raw Features\hfill{}(b) Discovered Attribute Predictions\hfill{}

\caption{\label{fig:clustering}t-SNE~\cite{Hinton2008} embedding of materials
from the raw feature space (a) and from our discovered attributes
(b). We embed a set of material image patches into 2D space via t-SNE
using raw features and predicted attribute probabilities as the input
space for the embeddings. Though t-SNE has been shown to perform well
in high-dimensional input spaces, it fails to separate material categories
from the raw feature space. Material categories are, however, clearly
more separable with our attribute space.}
\end{figure*}

We have observed that, similar to category-attribute associations,
predicted probabilities for known material traits are also Beta-distributed.
Local image regions exhibiting a trait will have uniformly high probability
for that trait, only decreasing around the trait region edges. We
constrain the predicted probabilities such that they are Beta-distributed.
Using the formulation discussed in Section~\ref{subsec:Defining-the-Material},
we again minimize a KL-divergence of a kernel density estimate:
\begin{equation}
\kappa\left(\mathbf{X};\mathbf{\Theta}\right)=\sum_{p\in P}\beta\left(p;a,b\right)\ln\left(\frac{\beta\left(p;a,b\right)}{q\left(p;f\left(\mathbf{X};\mathbf{\Theta}\right)\right)}\right)\text{,}\label{eq:kl_kde_classifier}
\end{equation}
where $f\left(\mathbf{X};\mathbf{\Theta}\right)$ represents the $N\times M$
matrix of attribute probability predictions for the training dataset,
and $q,a,b$ are defined as in Equation~\ref{eq:KL_KDE}.

One of the goals for our attribute representation is to discover attributes
that allow for material classification. If this were our only goal,
we could simply maximize the distance between the predicted attributes
for all pairs of different material categories. This would conflict
with our goal of preserving human perception, as material categories
do not always exhibit different appearances. We instead modify this
separation by weighting each component of the distance based on the
values in the category-attribute matrix:

\vspace{-15pt}
\begin{eqnarray}
 &  & \pi\left(\mathbf{X};\mathbf{A},\mathbf{\Theta}\right)=\sum_{i,j\in N|c_{i}\neq c_{j}}\mathbf{p}_{ij}^{\mathrm{T}}\mathbf{p}_{ij}\label{eq:pairwise_term}\\
 &  & \mathbf{p}_{ij}=\left(2\left|\mathbf{a}_{c_{i}}-\mathbf{a}_{c_{j}}\right|-1\right)\left(f\left(\mathbf{x}_{i};\mathbf{\Theta}\right)-f\left(\mathbf{x}_{j};\mathbf{\Theta}\right)\right)\text{.}\nonumber 
\end{eqnarray}
This separates the material categories in attribute space only when
the attributes dictate that there is a perceptual difference.

\subsection{\label{sec:Analysis-of-Discovered}Analysis of Discovered Attributes}

\begin{figure}
\begin{centering}
\includegraphics[width=0.19\columnwidth]{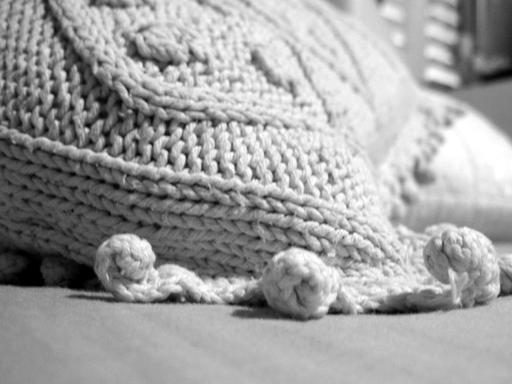}
\includegraphics[width=0.19\columnwidth]{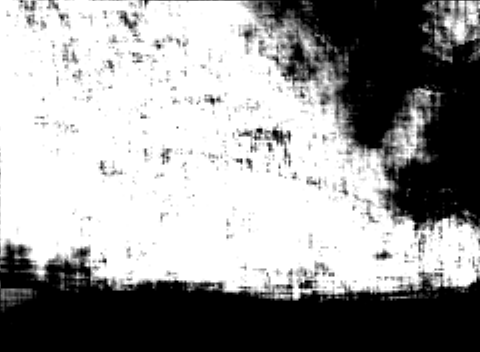}
\includegraphics[width=0.19\columnwidth]{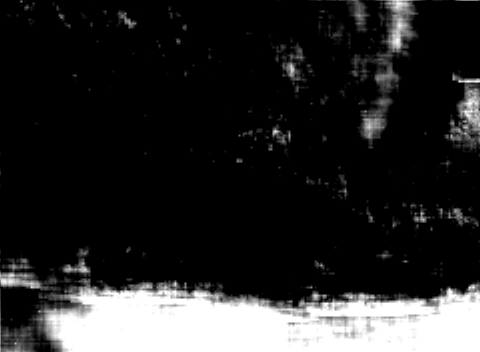}
\includegraphics[width=0.19\columnwidth]{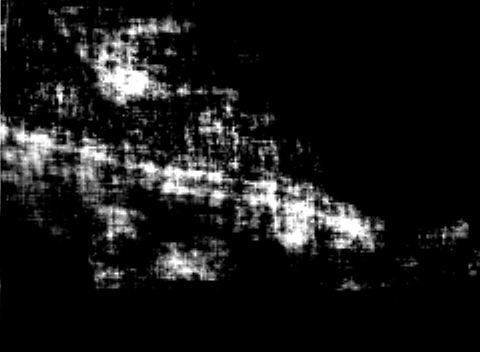}
\includegraphics[width=0.19\columnwidth]{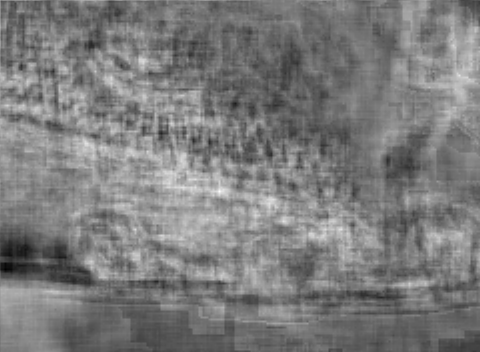}
\par\end{centering}
\vspace{2pt}
\begin{centering}
\includegraphics[width=0.19\columnwidth]{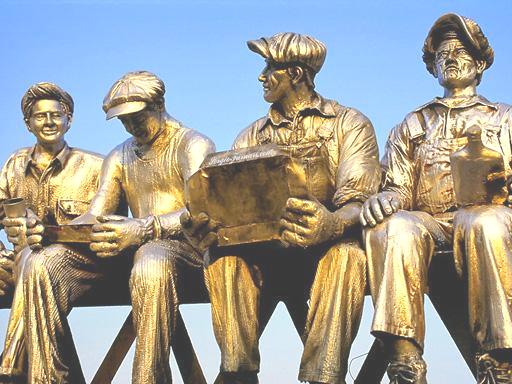}
\includegraphics[width=0.19\columnwidth]{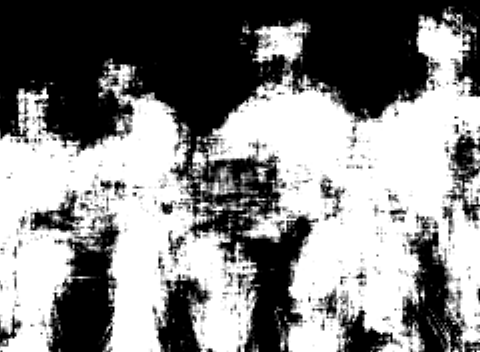}
\includegraphics[width=0.19\columnwidth]{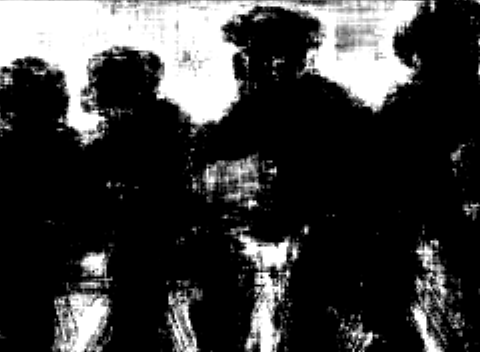}
\includegraphics[width=0.19\columnwidth]{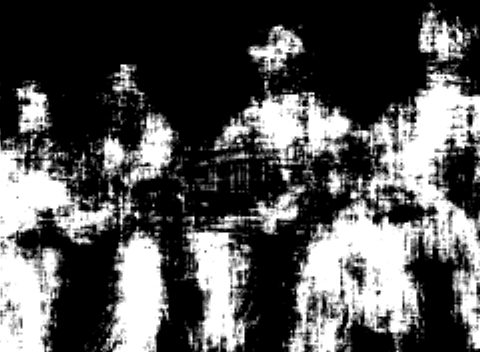}
\includegraphics[width=0.19\columnwidth]{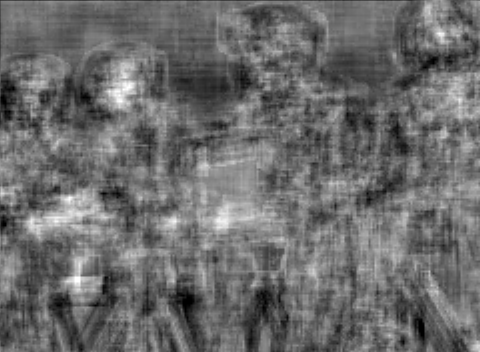}
\par\end{centering}
\vspace{2.5pt}
\begin{centering}
\includegraphics[width=0.19\columnwidth]{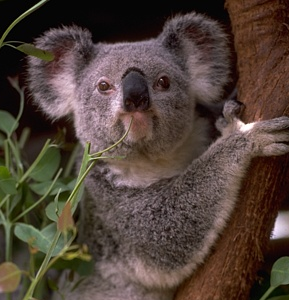} \includegraphics[width=0.19\columnwidth]{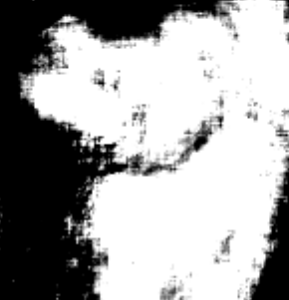}
\includegraphics[width=0.19\columnwidth]{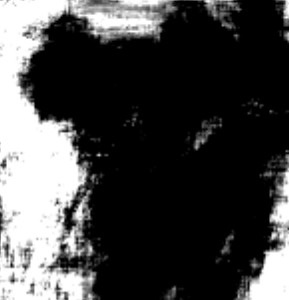}
\includegraphics[width=0.19\columnwidth]{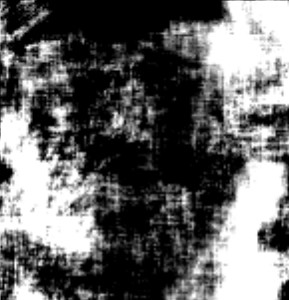}
\includegraphics[width=0.19\columnwidth]{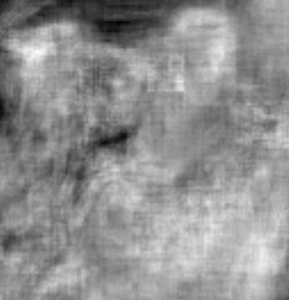}
\par\end{centering}
\begin{centering}
{\footnotesize{}\hfill{}\hspace{5pt}Input\hspace{5pt}\hfill{}Attribute
1\hfill{}Attribute 2\hfill{}Attribute 3\hspace{5pt}\hfill{}Random\hfill{}}
\par\end{centering}{\footnotesize \par}
\caption{\label{fig:Per-pixel-learned-attribute}Per-pixel discovered attribute
probabilities for four attributes (one per column). These images show
that the discovered attributes exhibit patterns similar to those of
known material traits. The first attribute, for example, appears consistently
within the woven hat and the koala; the second attribute tends to
indicate smooth regions. The third attribute shows we are discovering
attributes that can appear both sparsely and densely in an image,
depending on the context. These are all properties shared with visual
material traits. Attributes from a random $\mathbf{A}$ do not exhibit
any of these properties.}
\end{figure}

To analyze the properties of attributes discovered by our framework,
we follow the procedures outlined above to collect annotations and
discover a set of attributes. Since both learning steps involve minimization
of a non-linear, non-convex function, we rely on existing optimization
tools\footnote{Specifically, L-BFGS with box constraints for $\mathbf{A}$ and stochastic
gradient descent for $\mathbf{\Theta}$.} to find suitable estimates. As a raw feature set, we use the local
features we developed for material trait recognition (Section~\ref{sec:Visual-Material-Traits}).

If our attributes described a space that successfully separates material
categories, we would expect categories to form clusters in the attribute
space. To verify this, we compute a 2D embedding of a set of labeled
image patches. For the embedding, we use the t-SNE method of van~der~Maaten
and Hinton~\cite{Hinton2008}. t-SNE attempts to generate an embedding
that matches the distributions of neighboring points in the high-
and low-dimensional spaces. In \figword\ref{fig:clustering}, we
represent image patches by their raw feature vectors (a) and predicted
attribute probability vectors (b), and compare the 2D embeddings resulting
from each. Material categories are separated much more clearly in
our attribute space than in the raw feature space.

Part of the usefulness of visual material traits, as we have shown
above, is derived from the fact that they each represent a particular
intuitive visual material property. This is evident in the spatial
sparsity pattern of the traits, specifically the fact that they appear
in regions and not randomly within an image. Traits such as ``shiny''
are highly localized, while others such as ``woven'' or ``smooth''
exist as coherent regions within a particular material instance. \figword\ref{fig:Per-pixel-learned-attribute}
shows examples of per-pixel attribute probabilities predicted from
our discovered attribute classifiers. The attributes exhibit both
sparse and dense spatial patterns that are consistent within local
regions. Dense attributes generally correspond with smooth image regions.
Sparse attributes often indicate localized surface features such as
specific texture patterns.

Randomized features have been shown to provide a viable representation
for classification tasks~\cite{Rahimi2008}. Despite this, we expect
that such features would not be likely to encode the same perceptual
properties as our attributes. We can demonstrate this by replacing
our perceptually-derived attribute matrix $\mathbf{A}$ with randomized
matrix of the same shape. The last column in \figword\ref{fig:Per-pixel-learned-attribute}
shows typical results for such a matrix. Unlike attributes based on
human perception, these random attributes do not exhibit any of the
desired perceptual properties like spatial consistency.

\begin{figure}
\begin{centering}
\includegraphics[width=0.95\columnwidth]{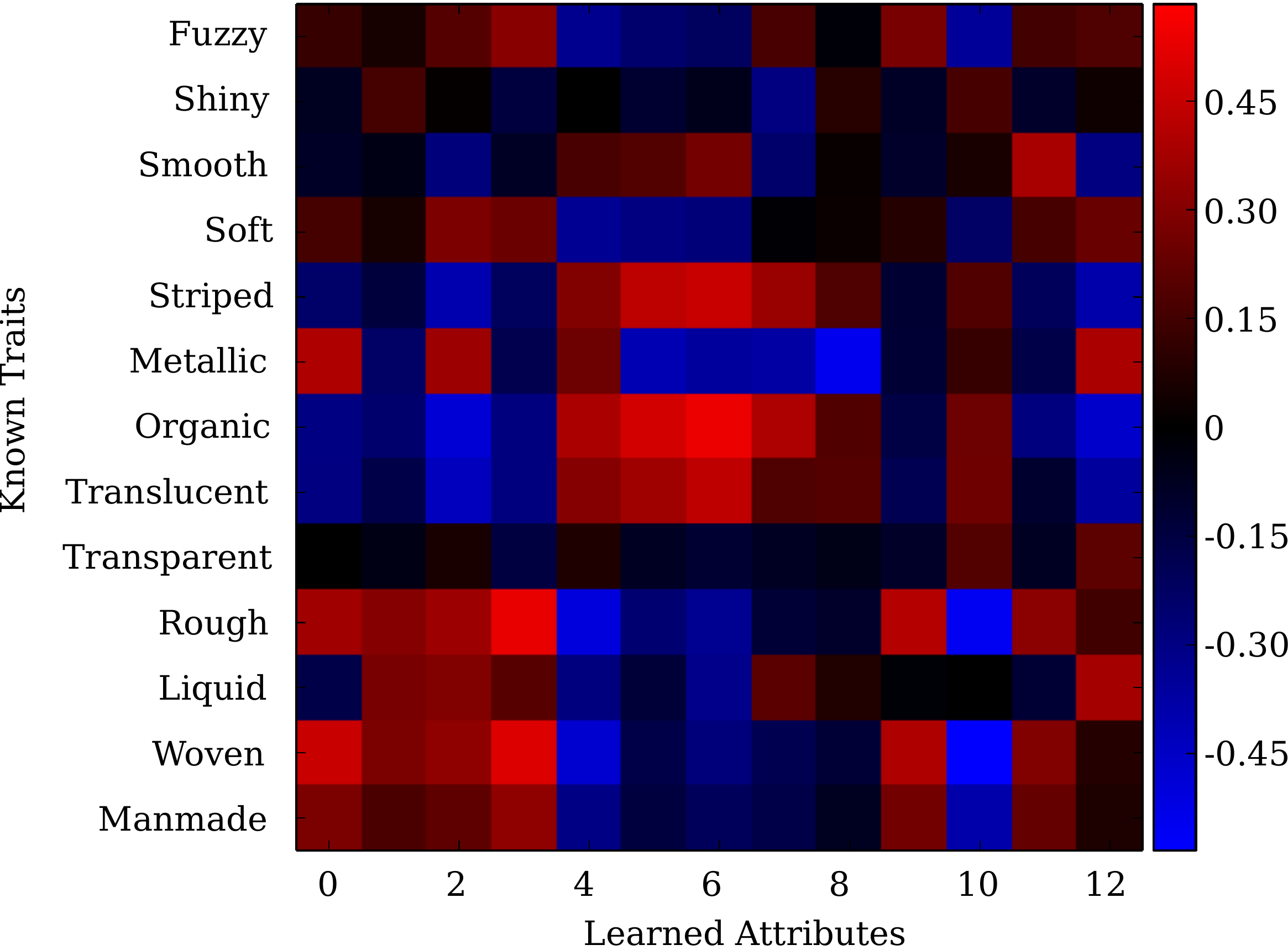}
\par\end{centering}
\caption{\label{fig:Correlation-between-known}Correlation between discovered
attribute predictions and material traits. Groups of attributes can
collectively indicate the presence of a material trait. Metallic,
for example, correlates positively with attribute 0 and negatively
with attribute 8.}
\end{figure}

We aimed to discover attributes similar to the visual material traits
that underlie human perception. We thus expect that the discovered
attributes exhibit a correlation with known traits. \figword\ref{fig:Correlation-between-known}
shows the correlation between 13 discovered attributes and 13 known
material traits using attributes predicted on labeled material trait
image patches. Collectively, we can indeed describe material traits
using the discovered attributes. Visually similar traits, such as
rough and woven, show similar correlations with the attributes. Discovered
attributes are also consistent with the semantic properties of material
traits. Rough and smooth are mutually exclusive traits, and we see
that discovered attributes that positively correlate with smooth do
not generally correlate with rough.

We quantitatively evaluate the discovered attributes using logic regression~\cite{Ruczinski2003}.
Given a set of image patches with known traits, we predict our discovered
attributes as binary values for use as input variables in a logic
regression model for material traits. Logic regression from 30 attributes
alone (no other features) achieves comparable accuracy to our trait-based
method and its complex feature set. These results show that the discovered
attributes do collectively encode intuitive visual material properties.
Further details and examples of logic regression to predict material
traits can be found in Sec.~\ref{subsec:Properties-of-the}.

\subsection{\label{sec:From-Discovered-Attributes}From Discovered Attributes
to Materials}

Seeing that discovered attributes encode visual material properties,
we would expect them to also serve as an intermediate representation
for material category recognition. To test this, we follow our local
material recognition procedure (Section~\ref{sec:Visual-Material-Traits}),
substituting our discovered attributes in place of labeled material
traits. We compute the histograms of these predicted probabilities
across the material region and use them as input for a histogram kernel
SVM. As we focus on local attributes, these previous local results
(and those of Sharan~\etal\cite{Sharan2013} on scrambled images)
serve as the correct baseline.

To compare with our previous results using material traits, we compute
average material recognition accuracy on the Flickr Materials Database
(FMD). All results are computed using $M=30$ discovered attributes
and 5-fold cross-validation unless otherwise specified.

Our attributes achieve an average accuracy of 48.9\% ($\sigma=1.2\%$)
on FMD images using only local information. This is comparable to
our results and those of Sharan~\etal\cite{Sharan2013} (using only
local information) even though we are discovering attributes using
only weak supervision.

Looking at individual class recognition rates, metal is the most challenging
category to identify while foliage is the most accurately recognized.
This follows from the results of our measurements of human perception,
as annotators consistently found that foliage image patches looked
different from all other material categories. Metal is particularly
difficult to recognize locally as its appearance depends strongly
on the appearance of the surround environment (due to specular reflection).

\begin{figure}
\begin{centering}
\includegraphics[width=0.95\columnwidth]{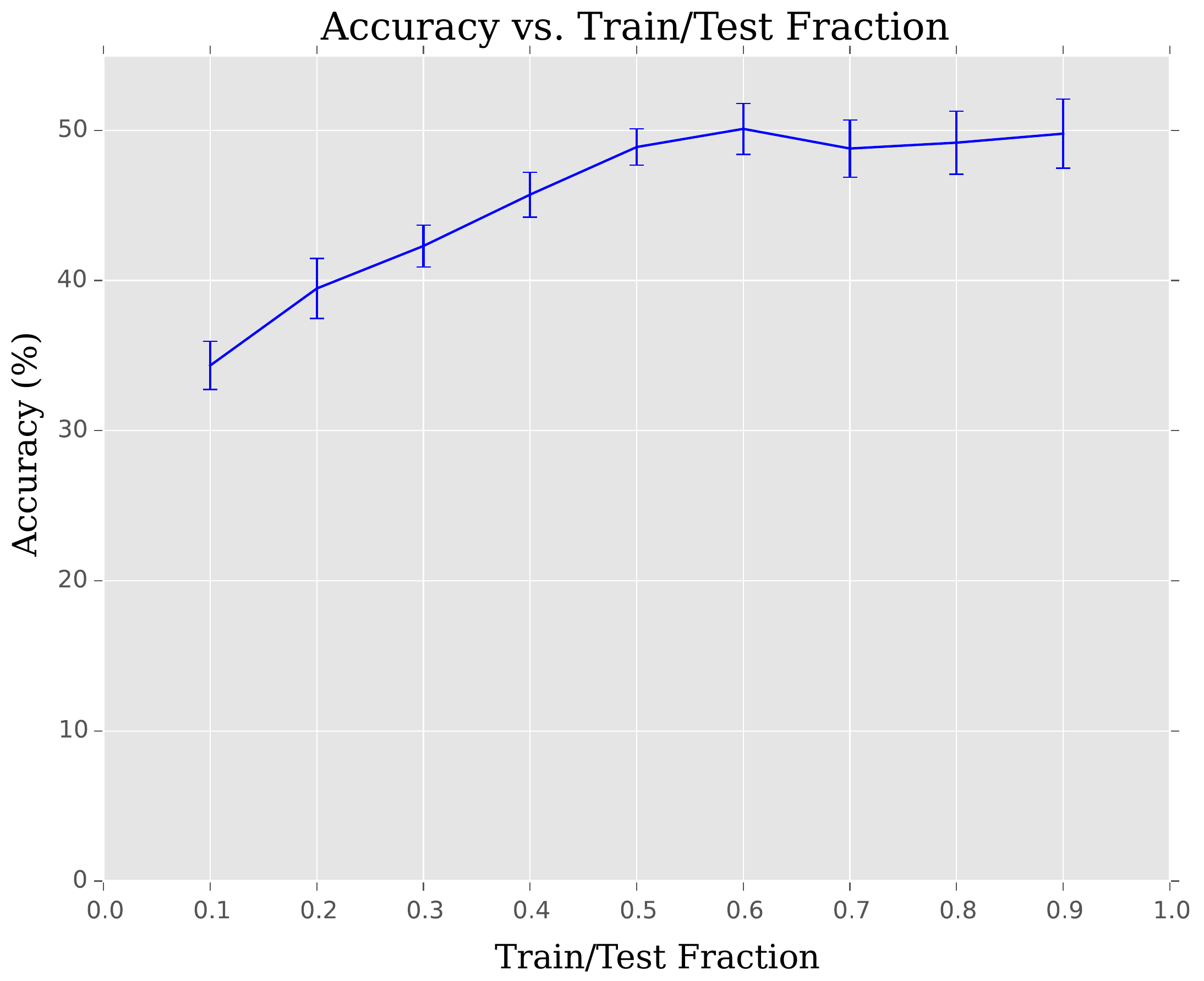}
\par\end{centering}
\caption{\label{fig:Accuracy-vs.-training}Accuracy vs. training set size.
The line is the average of 3 random splits. Error bars indicate the
minimum and maximum of the splits. Accuracy does not continue to increase
as we use larger training datasets. This shows that we have successfully
extracted as much local information as possible from human perception.}
\end{figure}

\figword\ref{fig:Accuracy-vs.-training} shows that accuracy reaches
a plateau as the training dataset size increases. We also compute
accuracy for varying values of $M$ and find that past $M=30$, there
is little (\textless{}0.1\%) gain in accuracy from additional attributes.
These results indicate that we are in fact extracting as much perceptual
material information as we can from the available data.

\section{Perceptual Material Attributes in Convolutional Neural Networks}

We have shown that we may use visual material traits to enable local
material recognition, and we may further scale this attribute-based
recognition process by automatically discovering perceptual material
attributes. Our previous methods consider attributes separately from
category recognition. The attributes are used solely as an intermediate
representation for material categories. Similarly for conventional
object and scene recognition, attributes like ``sunset'' or ``natural,''
have also been extracted for use as independent features. Shankar~\etal\cite{Shankar2015}
generate pseudo-labels to improve the attribute prediction accuracy
of a Convolutional Neural Network, and Zhou~\etal\cite{Zhou2015}
discover concepts from weakly-supervised image data. In both cases,
the attributes are considered on their own, not within the context
of higher-level categories. In object and scene recognition, however,
recent work shows that semantic attributes seem to arise in networks
that are trained end-to-end for category recognition~\cite{Zhou2015_2}.

We would like to take advantage of the benefits of end-to-end learning
to incorporate automatically-discovered attributes with material recognition
in one seamless process. Material attribute recognition, however,
is not easily scalable. In the past we relied on semantic attributes,
such as ``shiny'' or ``fuzzy'', that needed careful annotation
by a consistent annotator as their appearance may not be readily agreed
upon. We addressed the difficulty in annotation scaling by automatically
discovering perceptual material attributes from weak supervision.
The training process for this method does not, however, scale well
to large datasets. To address this, we propose a novel CNN architecture
that recognizes materials from small local image patches while producing
perceptual material attributes as an auxiliary output. We also introduce
a novel material database with material categories drawn from a materials-science-based
category hierarchy.

\subsection{\label{subsec:CNN-Attribute-Presence}Finding Material Attributes
in a Material Recognition CNN}

Hiramatsu~\etal\cite{Hiramatsu2011} have shown that perceptual
attributes form an integral component of the human material recognition
process. They found that during the process of material recognition,
we form a perceptual representation of materials analogous to the
intermediate representation provided by named material traits or automatically-discovered
material attributes. The key difference is that the perceptual representation
in human material recognition forms as an integral part of the recognition
process.

Our goal is to discover unnamed material attributes while performing
material recognition in one end-to-end scalable process. Based on
correlations between Convolutional Neural Network (CNN) feature maps
and human visual system neural output discovered by Yamins~\etal\cite{Yamins2014},
a CNN architecture appears to be a very suitable framework in which
to discover attributes analogous to those in human material perception.
Their work focuses on object recognition, however, and does not extract
any attributes. In this case, our automatically-discovered unnamed
material attributes are particularly relevant. We derive a novel framework
to discover perceptual attributes similar to the ones we describe
in Section~\ref{sec:Material-Attribute-Discovery} inside a material
recognition CNN framework.

A simple experiment to verify the feasibility of perceptual attribute
discovery in a CNN trained to recognize materials would be to add
a layer at the top of the network, immediately before the final material
category probability softmax layer, and constrain this layer to output
the same attributes we previously discovered in a separate process.
If we could predict attributes from this layer without affecting the
material recognition accuracy, this would suggest that we could indeed
combine attribute discovery and material recognition. We implemented
this approach and found that while the material accuracy was unaffected,
the attribute predictions were less accurate than if the CNN was trained
solely to predict the attributes and not materials as well (mean average
error of 0.2 vs 0.08).

The key issue with the straightforward approach is that it is not
an entirely faithful model to the process described in~\cite{Hiramatsu2011}.
They note that the human neural representation of material categories
transitions from visual (raw image features) to perceptual (visual
properties like ``shiny'') in an hierarchical fashion. This implies,
in agreement with findings of Escorcia~\etal\cite{Escorcia2015},
that attributes require information from multiple levels of the material
recognition network. We show that this is indeed the case by successfully
discovering the attributes using input from multiple layers of the
material recognition network.

\subsection{\label{subsec:Combined-Attribute/Material-CNN}Material Attribute-Category
CNN}

\begin{figure}
\begin{centering}
\includegraphics[width=0.98\columnwidth]{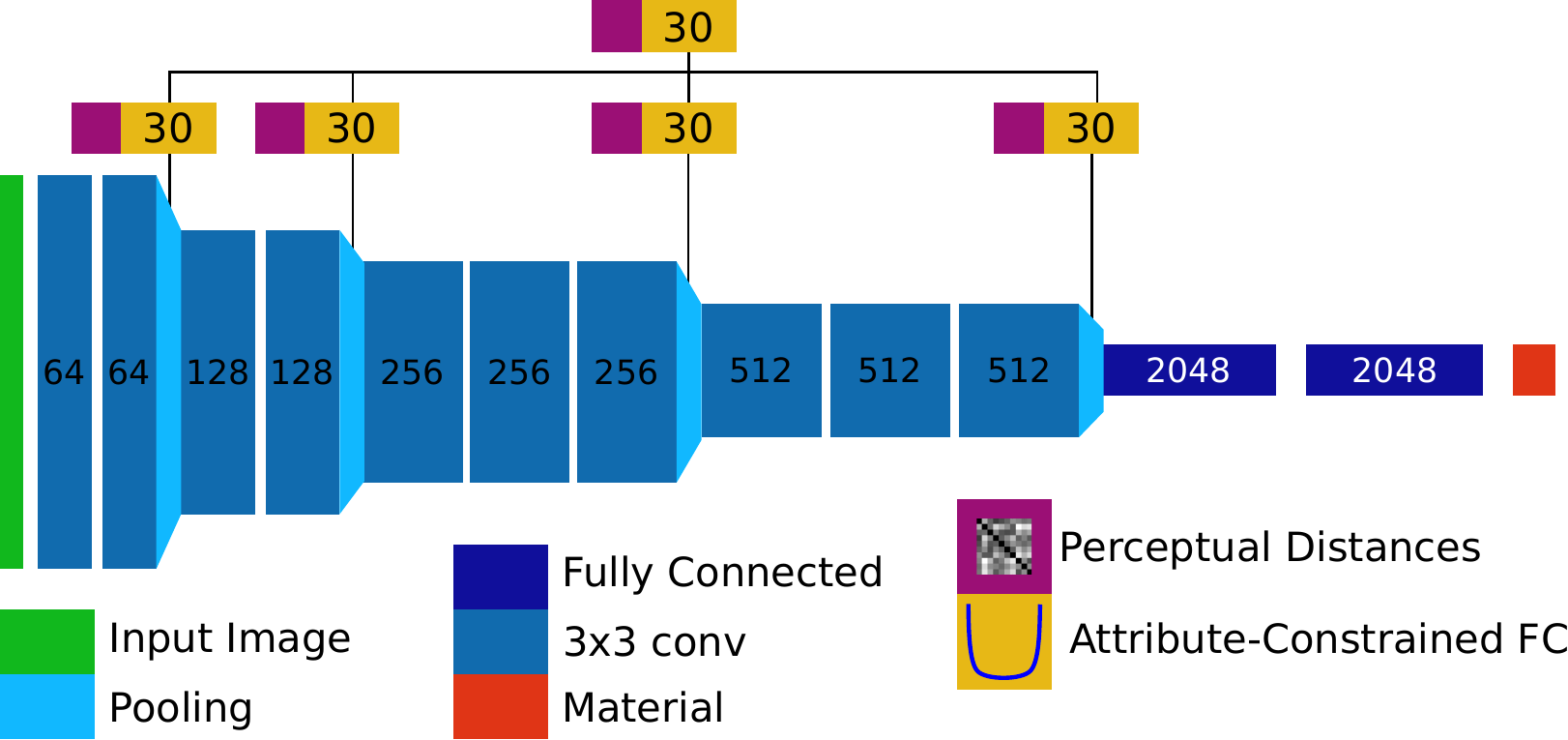}
\par\end{centering}
\caption{\label{fig:CNN-Architecture}Material Attribute-Category CNN (MAC-CNN)
Architecture: We introduce auxiliary fully-connected attribute layers
to each spatial pooling layer, and combine the per-layer predictions
into a final attribute output via an additional set of weights. The
loss functions attached to the attribute layers encourage the extraction
of attributes that match the human material representation encoded
in perceptual distances. The first set of attribute layers acts as
a set of weak learners to extract attributes wherever they are present.
The final layer combines them to form a single prediction.}
\end{figure}

We need a means of extracting attribute information at multiple levels
of the network. Simply combining all feature maps from all network
layers and using them to predict attributes would be computationally-impractical.
Rather than directly using all features at once, we augment an initial
CNN designed for material classification with a set of auxiliary fully-connected
layers attached to the spatial pooling layers. This allows the attribute
layers to use information from multiple levels of the network without
needing direct access to every feature map. We treat the additional
layers as a set of weak learners, each auxiliary layer discovering
the attributes available at the corresponding level of the network.
This is similar to the deep supervision of Lee~\etal\cite{Lee2015}.
Their auxiliary loss functions, however, simply propagate the same
classification targets (via SVM-like loss functions) to the lower
layers. Rather than propagating gradients, our attribute layers discover
perceptual material attributes.

For the auxiliary layer loss functions, we introduce a modified form
of the perceptual attribute loss function (Equation~\ref{eq:classifier_minimization})
to the outputs of each auxiliary fully-connected layer. Specifically,
assuming the output of a given pooling layer $i$ in the network for
image $j$ is $\mathbf{h}_{ij}$, and given categories $C,\,\left|C\right|=K$
and a set of sample points $P\in\left(0,1\right)$ for density estimation,
we add the following auxiliary loss functions:
\begin{equation}
u_{i}=\frac{1}{K}\sum_{k\in C}\left\Vert \mathbf{a}_{k}-\frac{1}{N_{k}}\sum_{j|c_{j}=k}f\left(\mathbf{W}_{i}^{\mathrm{T}}\mathbf{h}_{ij}+\mathbf{b}_{i}\right)\right\Vert _{1}\label{eq:unary}
\end{equation}
\begin{equation}
d_{i}=\sum_{p\in P}\beta\left(p;a,b\right)\ln\left(\frac{\beta\left(p;a,b\right)}{q\left(p;f\left(\mathbf{W}_{i}^{\mathrm{T}}\mathbf{h}_{ij}+\mathbf{b}_{i}\right)\right)}\right)\text{,}\label{eq:distr}
\end{equation}
where $f\left(x\right)=\min\left(\max\left(x,0\right),1\right)$ clamps
the outputs within $\left(0,1\right)$ to conform to attribute probabilities,
and weights $\mathbf{W}_{i},\mathbf{b}_{i}$ represent the auxiliary
fully-connected layers we add to the network. $\mathbf{a}_{k}$ represents
a row in the category-attribute mapping matrix derived as in Section~\ref{subsec:Defining-the-Material}.
Equation~\ref{eq:unary} causes the attribute layer to discover attributes
which match the perceptual distances measured from human annotations.
As certain attributes are expected to appear at different levels of
the network, some layers will be unable to extract them. This implies
that their error should be sparse, either predicting an attribute
well or not at all. For this reason we use an L1 error norm. Equation~\ref{eq:distr},
applied only to the final attribute layer, encourages the distribution
of the attributes to match those of known semantic material traits.
It takes the form of a KL-divergence between a Beta distribution (empirically
observed to match the distribution of semantic attribute probabilities),
and a Kernel Density Estimate $q\left(\cdot\right)$ of the extracted
attribute probability density sampled at points $p\in P$.

The reference network we modify is based on the high-performing VGG-16
network of Simonyan and Zisserman~\cite{Simonyan2015}. We use their
trained convolutional weights as initialization where applicable,
and add new fully-connected layers for material classification. \figword\ref{fig:CNN-Architecture}
shows our architecture for material attribute discovery and category
recognition. We refer to this network as the Material Attribute-Category
CNN (MAC-CNN).

\section{\label{sec:Constructing-a-Local-Material-Database}Local Material
Database}

In order to train the category recognition portion of the MAC-CNN,
we need a suitable dataset, and we find existing material databases
lacking in a few key areas. Previous material recognition datasets
\cite{Sharan2013,Bell2013,Bell2015} have relied on ad-hoc choices
regarding the selection and granularity of material categories. When
patches are involved, as in~\cite{Bell2015}, the patches can be
as large as 24\% of the image size surrounding a single pixel identified
as corresponding to a material. These patches are large enough to
include entire objects. These issues make it difficult to separate
challenges inherent to material recognition from those related to
general recognition tasks. Materials, as with their visual attributes,
are inherently local properties. While knowledge of what object the
material composes may help recognize the material, the material is
not the object. We also find that image diversity is still lacking
in these modern datasets. For these reasons, we introduce a new local
material recognition dataset to support the experiments in this paper.

\subsection{\label{subsec:Material-Hierarchy}Material Category Hierarchy}

\begin{figure}
\begin{centering}
\includegraphics[width=1\columnwidth]{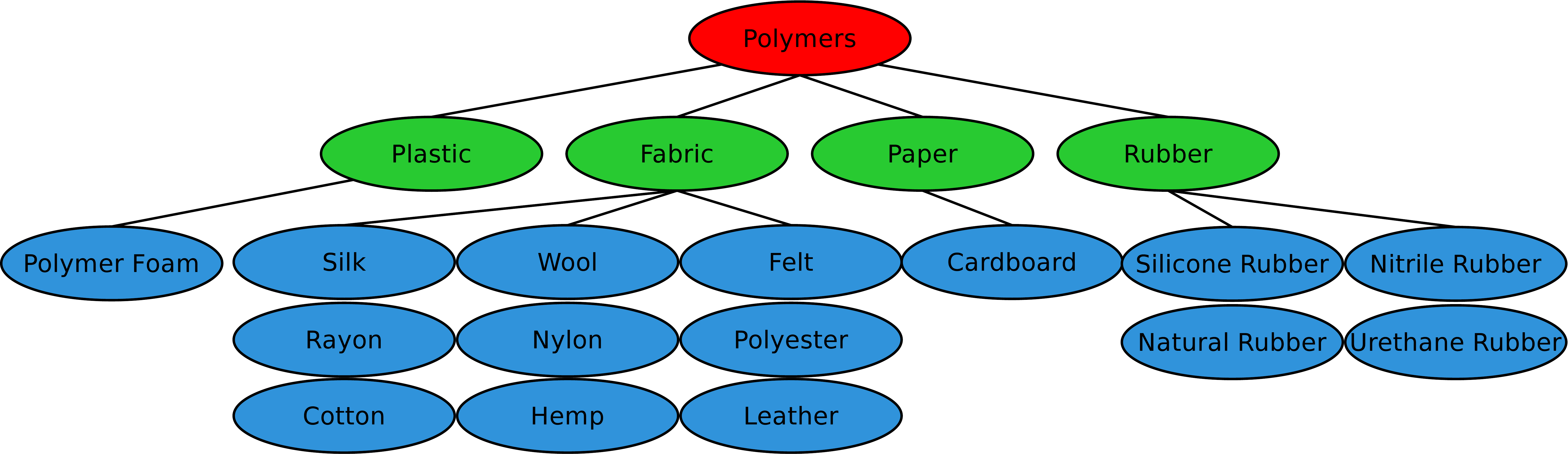}
\par\end{centering}
\caption{\label{fig:hierarchy}One tree in our material category hierarchy.
Categories at the top level separate materials with notable differences
in physical properties. Mid-level categories are visually distinct.
The lowest level of categories are fine-grained and may require both
physical and visual properties and expert knowledge to distinguish
them.}
\end{figure}

Material categories in existing datasets have not been carefully selected.
Examples of this appear in the recent work of Bell~\etal\cite{Bell2015},
where proposed material categories include ``mirror'' and ``carpet''
(among others). These are in fact objects, and their annotations reflect
this. Their categories also confuse materials and their visual properties,
for example, separating ``stone'' from ``polished stone''. To
address the issue of material category definition, we propose a set
of material categories derived from a materials science taxonomy.
Additionally, we create a hierarchy based on the generality of each
material family. \figword\ref{fig:hierarchy} shows an example of
one tree of the hierarchy.

Our hierarchy consists of a set of three-level material trees. The
highest level corresponds to major structural differences between
materials in the category. Metals are conductive, polymers are composed
of long chain molecules, ceramics have a crystalline structure, and
composites are fusions of materials either bonded together or in a
matrix. We define the mid-level (also referred to as entry-level~\cite{Ordonez2013})
categories as groups that separate materials based primarily on their
visual properties. Rubber and paper are flexible, for example, but
paper is generally matte and rubber exhibits little color variation.
The lowest level, fine-grained categories, can often only be distinguished
via a combination of physical and visual properties. Silver and steel,
for example, may be challenging to distinguish based solely on visual
information.

Such a hierarchy is sufficient to cover most natural and manmade materials.
In creating our hierarchy, however, we found that certain categories
that are in fact materials did not fit within the strict definitions
described above. For the sake of completeness, we make the conscious
decision to add these mid-level categories to our data collection
process. These categories are: food, water, and non-water liquids.
While food is both a material and an object, we rely on our annotation
process (Sec.~\ref{subsec:Crowdsourcing-Annotation}) to ensure we
obtain examples of the former and not the latter.

\subsection{\label{subsec:Crowdsourcing-Annotation}Data Collection and Annotation}

The mid-level set of categories forms the basis for a crowdsourced
annotation pipeline to obtain material regions from which we may extract
local material patches. We employ a multi-stage process to efficiently
extract both material presence and segmentation information for a
set of images.

The first stage asks annotators to identify materials present in the
image. Given a set of images with materials identified in each image,
the second stage presents annotators with a user interface that allows
them to draw multiple regions in an image. Each annotator is given
a single image-material pair and asked to mark regions where that
material is present. While not required, our interface allows users
to create and modify multiple disjoint regions in a single image.
Images undergo a final validation step to ensure no poorly drawn or
incorrect regions are included.

Each image in the first stage is shown to multiple annotators and
a consensus is taken to filter out unclear or incorrect identifications.
While sentinels and validation were not used to collect segmentations
in other datasets, ours is intended for local material recognition.
This implies that identified regions should contain only the material
of interest. During collection, annotators are given instructions
to keep regions within object boundaries, and we validate the final
image regions to insure this.

Image diversity is an issue present to varying degrees in current
material image datasets. The Flickr Materials Database (FMD)~\cite{Sharan2009}
contains images from Flickr which, due to the nature of the website,
are generally more artistic in nature. The OpenSurfaces and Materials
in Context datasets~\cite{Bell2013,Bell2015} attempt to address
this, but still draw from a limited variety of sources. We source
our images from multiple existing image datasets spanning the space
of indoor, outdoor, professional, and amateur photographs. We use
images from the PASCAL VOC database~\cite{pascal-voc-2008}, the
Microsoft COCO database~\cite{Lin2014}, the FMD~\cite{Sharan2009},
and the imagenet database~\cite{Russakovski2015}.

\begin{figure}
\begin{centering}
\includegraphics[width=0.98\columnwidth]{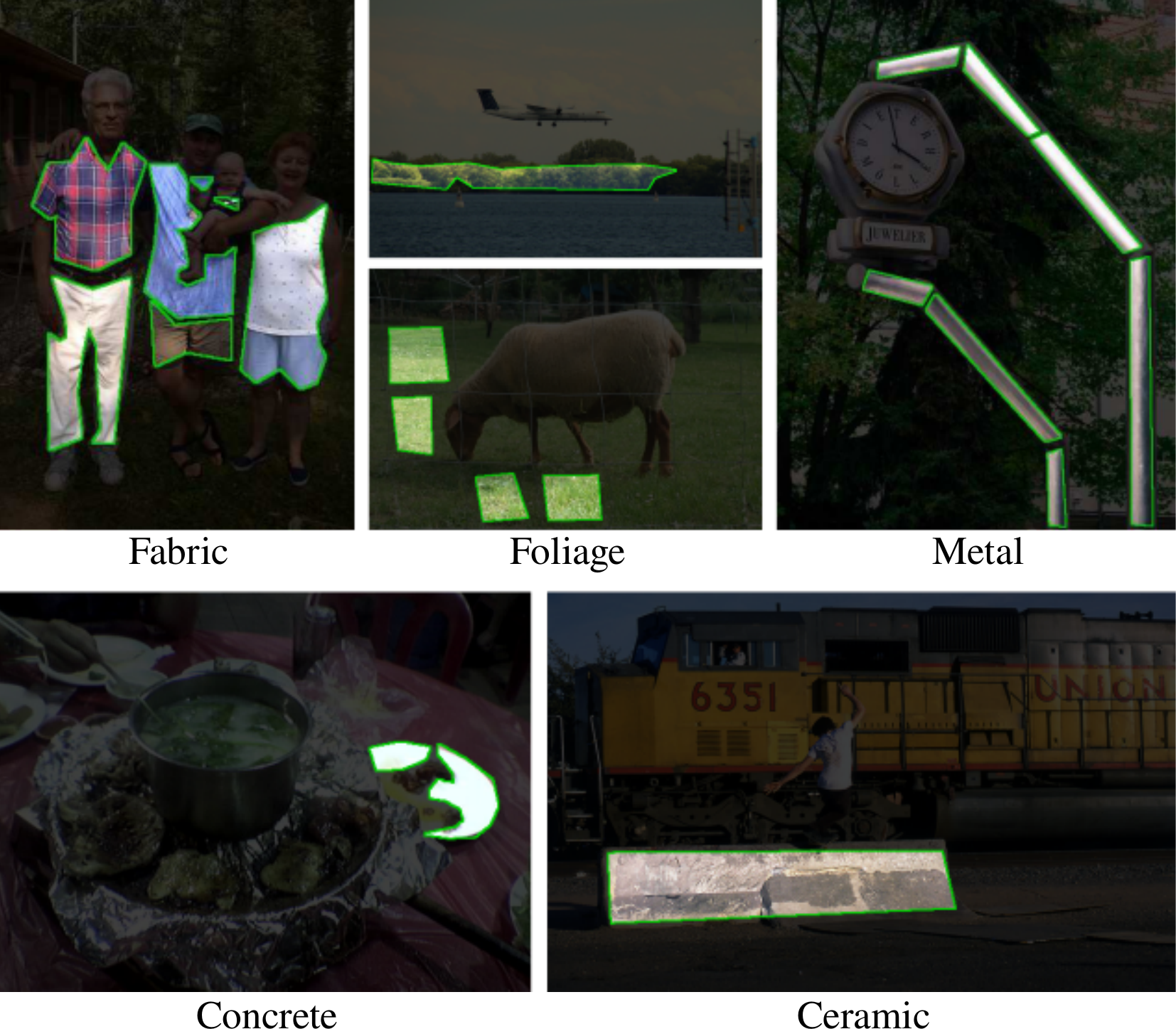}
\par\end{centering}
\caption{\label{fig:annotation-results}Example annotation results. Annotators
did not hesitate to take advantage of the ability to draw multiple
regions, and most understood the guidelines concerning regions crossing
object boundaries. As a result, we have a rich database of segmented
local material regions.}
\end{figure}

Examples in \figword\ref{fig:annotation-results} show that our annotation
pipeline successfully provides properly-segmented material regions
within many images. Many images also contain multiple regions. While
the level of detail for provided regions varies from simple polygons
to detailed material boundaries, the regions all contain single materials.

Our database currently contains 5845 images with carefully-segmented
material regions. For comparison, the MINC~\cite{Bell2015} database
does not provide any segmented material regions for training, and
only 1798 images for testing.

\section{\label{sec:Analysis-of-Perceptual}Visual Material Attributes Discovered
in the MAC-CNN}

To verify that the visual material attributes we seek are indeed present
in and can be extracted with our MAC-CNN, we augment our dataset with
annotations to compute the necessary perceptual distances described
in Section~\ref{subsec:Measuring-Perceptual-Distances}. Using our
dataset and these distances, we derive a category-attribute matrix
$\mathbf{A}$ and train an implementation of the MAC-CNN described
in Section~\ref{subsec:Combined-Attribute/Material-CNN}.

We train the network on \textasciitilde{}200,000 $48\times48$ image
patches extracted from segmented material regions. Optimization is
performed using mini-batch stochastic gradient descent with momentum.
The learning rate is decreased by a factor of 10 whenever the validation
error increases, until the learning rate falls below $1\times10^{-8}$.

\subsection{\label{subsec:Properties-of-the}Properties of MAC-CNN Visual Material
Attributes}

\begin{figure}
\begin{centering}
\includegraphics[height=0.2\paperheight]{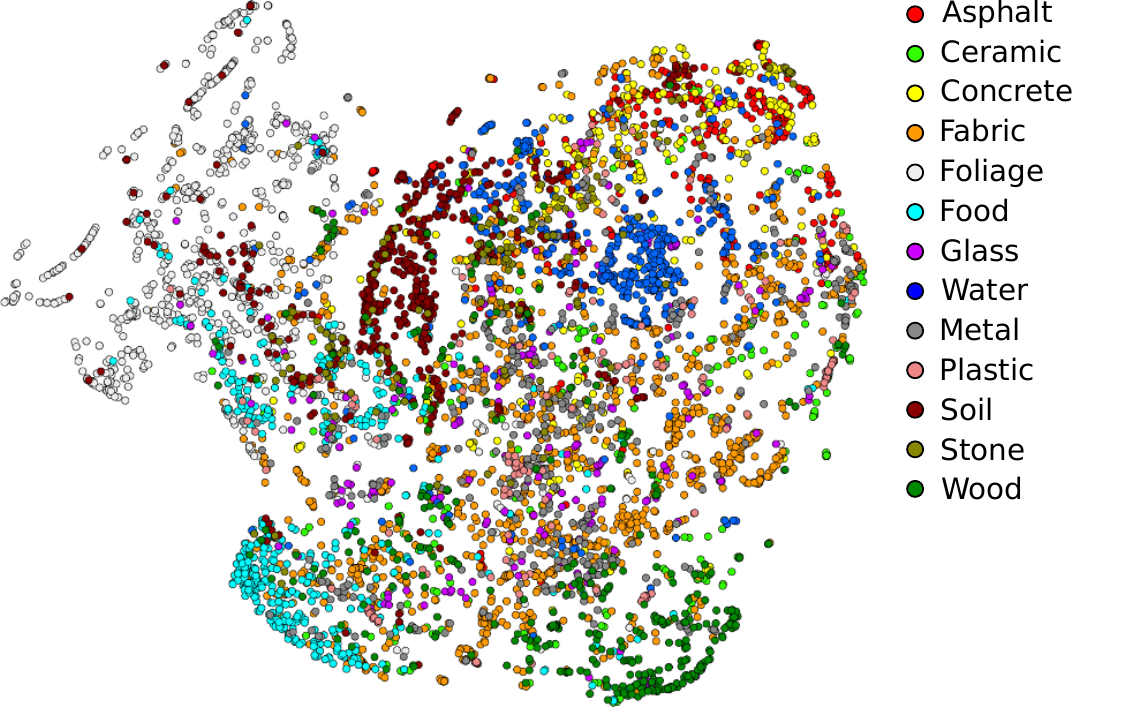}
\par\end{centering}
\caption{\label{fig:Attribute-Analysis}Attribute Space Embedding via t-SNE~\cite{Hinton2008}:
Many categories, such as water, food, foliage, soil, and wood, are
extremely well-separated in the attribute space. We find that this
separation corresponds with per-category accuracy: well-separated
categories are recognized more accurately. While other categories
do overlap to some extent, they still form separate regions in the
space.}
\end{figure}

\begin{figure}
\begin{centering}
\includegraphics[width=0.98\columnwidth]{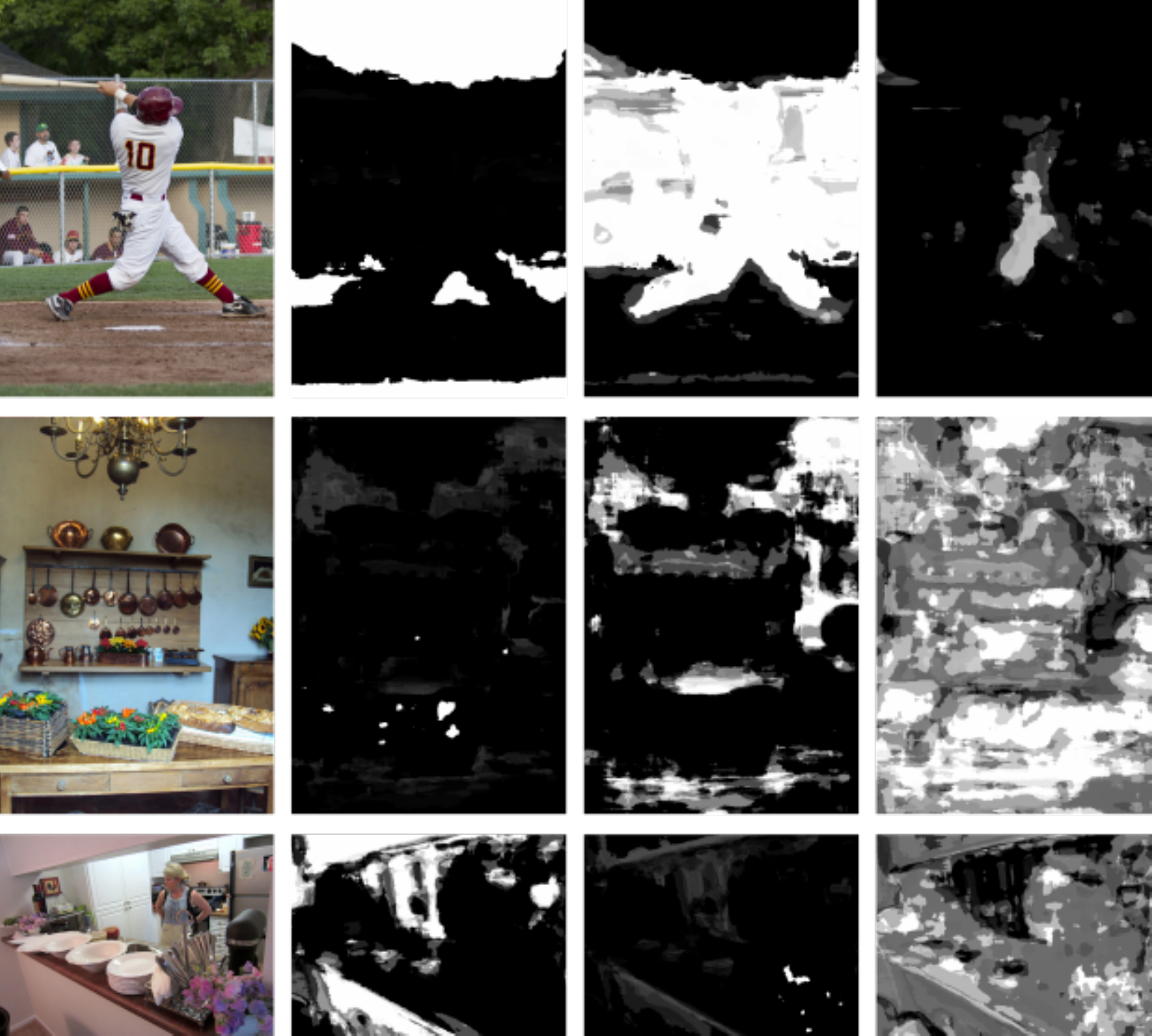}
\par\end{centering}
\caption{\label{fig:Per-Pixel-Attribute-Probabilitie}The attributes from the
MAC-CNN (visualized to the right of an input image above) exhibit
the same properties as our the attributes we previously discovered
independent of material recognition. The attributes form clearly delineated
regions, similar to semantic attributes, and their distributions match
as well.}
\end{figure}

\begin{figure}
\begin{centering}
\begin{minipage}[b][1\totalheight][t]{0.45\columnwidth}%
\begin{center}
\vspace{20pt}
\includegraphics[width=0.48\columnwidth]{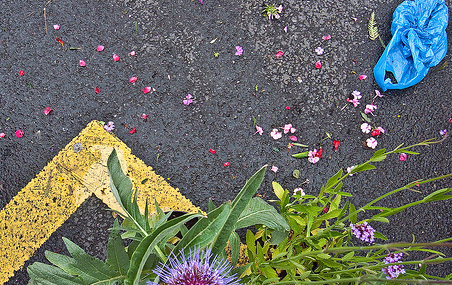}
\includegraphics[width=0.48\columnwidth]{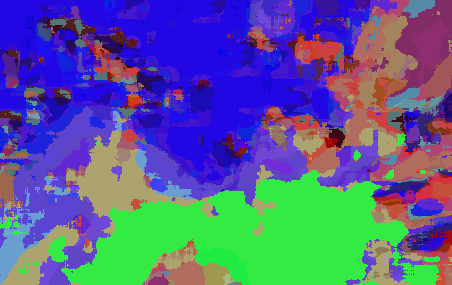}
\par\end{center}
\vspace{-15pt}
\begin{center}
\colorsquare[red]{4pt}\textsf{\scriptsize{} Manmade\hspace{5pt}}\colorsquare[green]{4pt}\textsf{\scriptsize{}
Organic\hspace{5pt}}\colorsquare[blue]{4pt}\textsf{\scriptsize{}
Rough}
\par\end{center}{\scriptsize \par}
\begin{center}
\includegraphics[width=0.48\columnwidth]{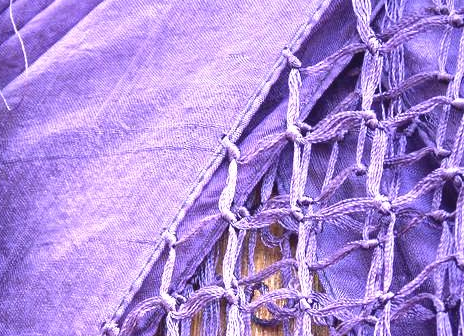}
\includegraphics[width=0.48\columnwidth]{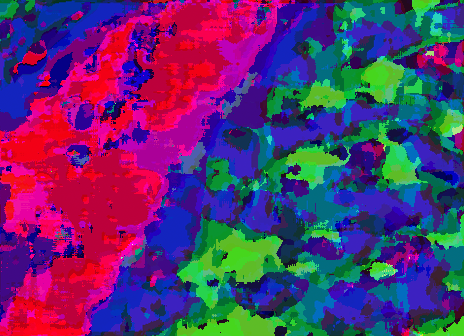}\vspace{-10pt}
\par\end{center}
\begin{center}
\colorsquare[red]{4pt}\textsf{\scriptsize{} Smooth\hspace{5pt}}\colorsquare[green]{4pt}\textsf{\scriptsize{}
Striped\hspace{5pt}}\colorsquare[blue]{4pt}\textsf{\scriptsize{}
Soft}
\par\end{center}{\scriptsize \par}%
\end{minipage} \hspace{5pt}%
\begin{minipage}[t]{0.49\columnwidth}%
\begin{center}
\vspace{-120pt}
\includegraphics[width=0.48\columnwidth]{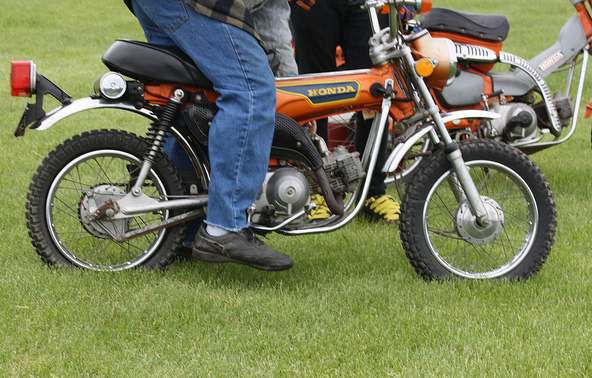}
\includegraphics[width=0.48\columnwidth]{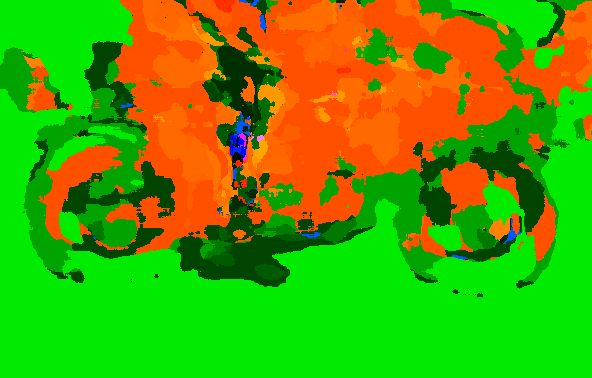}\vspace{-5pt}
\par\end{center}
\begin{center}
\colorsquare[red]{4pt}\textsf{\scriptsize{} Metallic\hspace{5pt}}\colorsquare[green]{4pt}\textsf{\scriptsize{}
Organic\hspace{5pt}}\colorsquare[blue]{4pt}\textsf{\scriptsize{}
Smooth}
\par\end{center}{\scriptsize \par}
\vspace{-5pt}
\begin{center}
\includegraphics[width=0.49\columnwidth]{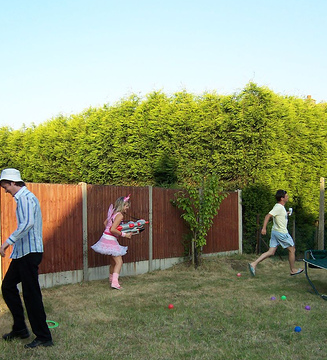}
\includegraphics[width=0.49\columnwidth]{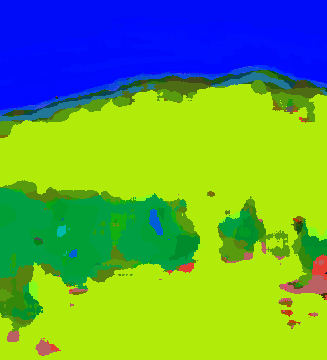}
\par\end{center}
\vspace{-10pt}
\begin{center}
\colorsquare[red]{4pt}\textsf{\scriptsize{} Fuzzy\hspace{5pt}}\colorsquare[green]{4pt}\textsf{\scriptsize{}
Organic\hspace{5pt}}\colorsquare[blue]{4pt}\textsf{\scriptsize{}
Smooth}
\par\end{center}{\scriptsize \par}%
\end{minipage}
\par\end{centering}
\caption{\label{fig:traits-from-attrs}By performing logic regression from
our MAC-CNN extracted attributes to semantic material traits, we are
able to extract semantic information from our non-semantic attributes.
We can apply logic regression to material attribute predictions on
patches in a sliding window to obtain per-pixel semantic material
trait information. The per-pixel trait predictions show crisp regions
that correspond well with their associated semantic traits. Traits
are independent, and thus the maps contain mixed colors. Fuzzy and
organic in the lower right image, for example, creates a yellow tint.}
\end{figure}

We examine the properties of our visual material attributes by visualizing
how they separate materials, computing per-pixel attribute maps to
verify that the attributes are being recognized consistently, and
linking the non-semantic attributes with known semantic material traits
(``fuzzy'', ``smooth'', etc...) to visualize semantic content.
Figs.~\ref{fig:Attribute-Analysis}, \ref{fig:Per-Pixel-Attribute-Probabilitie},
and~\ref{fig:traits-from-attrs} are generated using a test set of
held-out images.

A 2D embedding of material image patches shows that the  attributes
(\figword\ref{fig:Attribute-Analysis}) separate material categories.
A number of materials are almost completely distinct in the attribute
space, while a few form overlapping but still distinguishable regions.
Foliage, food, and water form particularly clear clusters. The quality
of the clusters correlates with category recognition accuracy, with
accurately-recognized categories forming well-separated clusters.

Visualizations of per-pixel attribute probabilities in \figword\ref{fig:Per-Pixel-Attribute-Probabilitie}
show that the attributes are spatially consistent. While overfitting
is difficult to measure for weakly-supervised attributes, we use spatial
consistency as a proxy. Spatial consistency is an indicator that the
attributes are not overly-sensitive to minute changes in local appearance,
something that would appear if overfitting were present. The attributes
exhibit correlation with the materials that induced them: attributes
with a strong presence in a material region in one image often appear
similarly in others. The visualizations also clearly show that the
attributes are representing more than trivial properties such as ``flat
color'' or ``bumpy texture''.

Logic regression~\cite{Ruczinski2003} is a method for building trees
that convert a set of boolean variables into a probability value via
logical operations (AND, OR, NOT). It is well-suited for collections
of binary attributes such as ours. Results of performing logic regression
(\figword\ref{fig:traits-from-attrs}) from extracted attribute predictions
to known semantic material traits (such as fuzzy, shiny, smooth etc...)
show that our MAC-CNN attributes encode material traits with the roughly
same average accuracy (77\%) as the our previous attributes. We may
also predict per-pixel trait probabilities in a sliding window fashion,
showing that the attributes are encoding both perceptual and semantic
material properties.

\subsection{\label{subsec:Local-Material-Recognition}Local Material Recognition}

\setlength\tabcolsep{1.25pt} \def\arraystretch{0.5}

Our results in Section~\ref{sec:Analysis-of-Perceptual} show that
we can successfully discover unnamed visual material attributes in
the combined material-attribute network. If these discovered attributes
were not complete, or if they were not faithfully representing material
properties, we would expect material recognition accuracy to improve
by removing the attribute discovery constraints. While the attribute
layers are auxiliary, they are connected to spatial pooling layers
at every level and thus the attribute constraints affect the entire
network. We in fact find that the average material category accuracy
does not change when the attribute layers are removed.

The average accuracy is 60.2\% across all categories. Foliage is the
most accurately recognized, consistent with past material recognition
results in which foliage is the most visually-distinct category. Paper
is the least well-recognized category. Unlike the artistic closeup
images of the FMD, many of the images in our database come from ordinary
images of scenes. Paper, in these situations, shares its appearance
with a number of other materials such as fabric.

It is important to note that we are recognizing materials directly
from single small image patches, with none of the region-based aggregation
or large patches used before and by other methods~\cite{Bell2015}.
This is a much more challenging task as the available information
is restricted. These restrictions are necessary, however, as using
large image patches (such as in the MINC database) would cause the
attributes we discover to implicitly depend on the objects present
in the patches and not simply the materials.

In this paper, our goal is not to introduce a novel material recognition
method but rather to investigate the recognition of material properties
from images. As such, the correct baseline for comparison is between
our previously discovered attributes and those we discover in the
MAC-CNN, not material recognition accuracy. Material recognition accuracy
is used simply to demonstrate that the discovery of visual material
attributes is compatible with material recognition in a single end-to-end
trainable network. As we have shown in our parallel work~\cite{Schwartz2017},
accurate material recognition requires the proper integration of local
and global context. Integrating our attribute discovery method with
a state-of-the-art material recognition framework that fuses local
and global context is one promising avenue for future work.

\begin{figure}
\begin{centering}
\noindent\begin{minipage}[b][1\totalheight][t]{1\columnwidth}%
\begin{center}
\begin{tabular}{ccc}
\colorsquare[red]{4pt}\hspace{3pt}\textsf{\scriptsize{}Soil\hspace{10pt}} & \colorsquare[green]{4pt}\hspace{3pt}\textsf{\scriptsize{}Foliage\hspace{10pt}} & \colorsquare[blue]{4pt}\hspace{3pt}\textsf{\scriptsize{}Fabric}\tabularnewline
\end{tabular}
\par\end{center}
\begin{center}
\includegraphics[width=0.48\columnwidth]{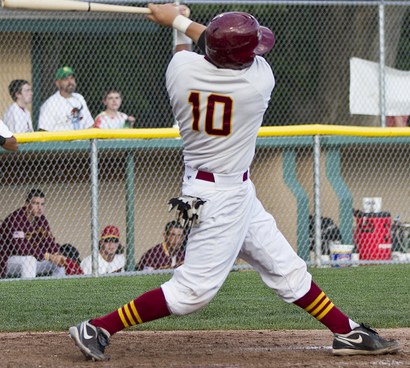}
\includegraphics[width=0.48\columnwidth]{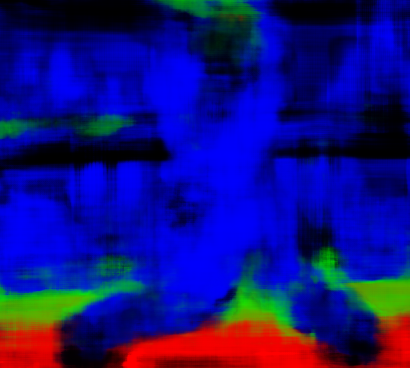}
\par\end{center}
\begin{center}
\includegraphics[width=0.48\columnwidth]{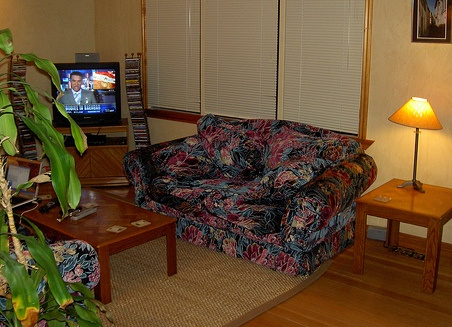}
\includegraphics[width=0.48\columnwidth]{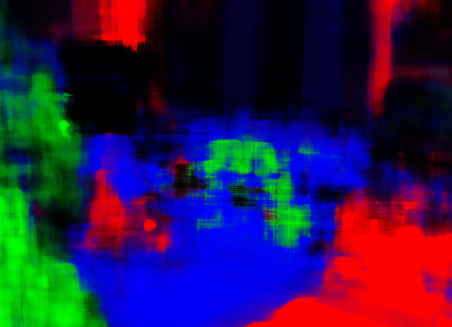}\\
\begin{tabular}{ccc}
\colorsquare[red]{4pt}\hspace{3pt}\textsf{\scriptsize{}Wood\hspace{10pt}} & \colorsquare[green]{4pt}\hspace{3pt}\textsf{\scriptsize{}Foliage\hspace{10pt}} & \colorsquare[blue]{4pt}\hspace{3pt}\textsf{\scriptsize{}Fabric}\tabularnewline
\end{tabular}
\par\end{center}%
\end{minipage}
\par\end{centering}
\caption{\label{fig:Material-Probability-Maps}Applying the MAC-CNN in a sliding-window
fashion leads to a set of material category probability maps. These
material maps show that we may obtain coherent regions using only
small local patches as input. The foliage predictions in the bottom
right image are reasonable, as the local appearance is indeed a flower.
In the upper right image, the local appearance of the fence resembles
lace (a fabric).}
\end{figure}

While a full dense per-pixel material segmentation framework is outside
the scope of this work, we are able to use the MAC-CNN to produce
per-pixel material probability predictions in a sliding window fashion.
Results in \figword\ref{fig:Material-Probability-Maps} show that
we may still generate reasonable material probability maps even from
purely local information.

\subsection{\label{sec:Novel-Material-Category}Novel Material Category Recognition}

One prominent application of attributes is in novel category recognition
tasks. Examples of these tasks include one-shot~\cite{Fei-Fei2006}
or zero-shot learning~\cite{Lampert2009}. Zero-shot learning allows
recognition of a novel category from a human-supplied list of applicable
semantic attributes. Since our attributes are non-semantic, zero-shot
learning is not applicable here. We may, however, investigate the
generalization of our attributes through a form of one-shot learning
in which we use image patches extracted from a small number of images
to learn a novel category.

To evaluate the application of visual material attributes for novel
category recognition, we train a set of attribute/material networks
on modified datasets each containing a single held-out category. No
examples of the held-out category are present during training. The
corresponding row of the category-attribute matrix is also removed.
The same number of attributes are defined based on the remaining categories.

For the novel category training, we use a balanced dataset consisting
of unseen examples of training categories and a matching number of
images from the held-out category. We also separate a number of images
of the held-out category as final testing samples. We train a simple
binary classifier (a linear SVM) to distinguish between the training
categories and the held-out category based on either their attribute
probabilities, material probabilities, or both, computed on patches
extracted from each input image. We measure the effectiveness of novel
category recognition by the fraction of final held-out category samples
properly identified as belonging to that category.

\begin{figure*}
\begin{centering}
\includegraphics[width=0.66\columnwidth]{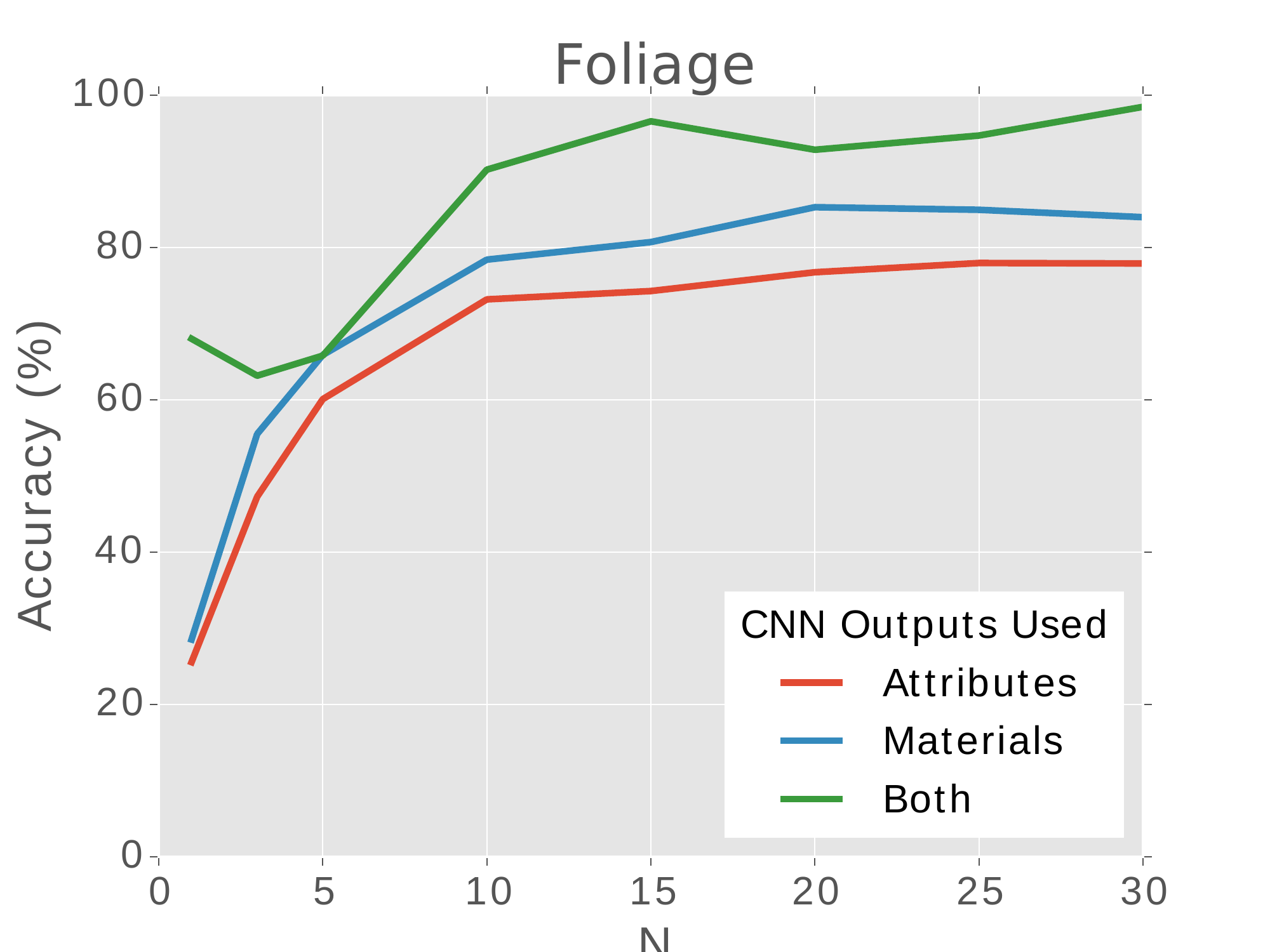}\includegraphics[width=0.66\columnwidth]{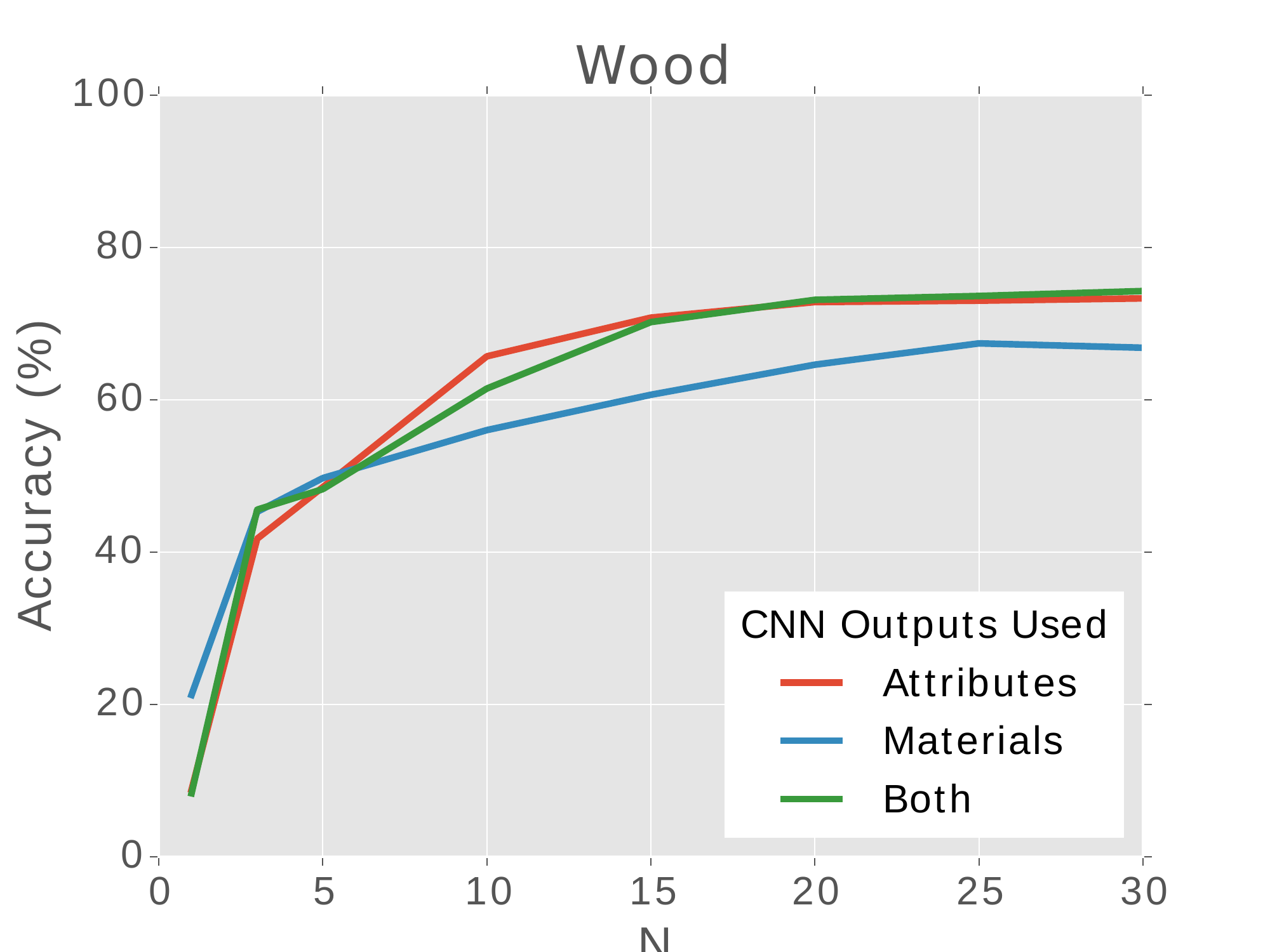}\includegraphics[width=0.66\columnwidth]{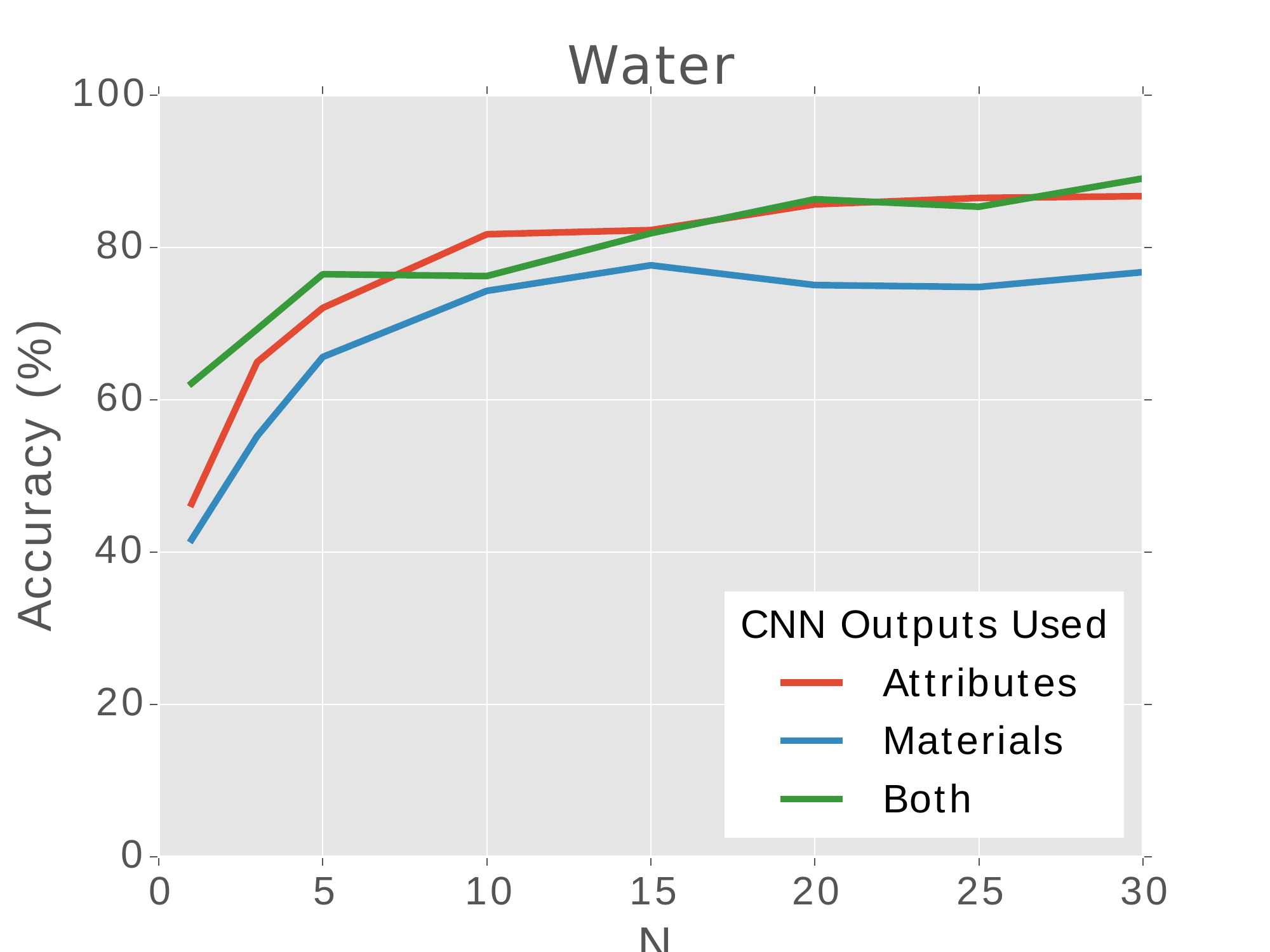}
\par\end{centering}
\begin{centering}
\hspace{-10pt}\includegraphics[width=0.61\columnwidth]{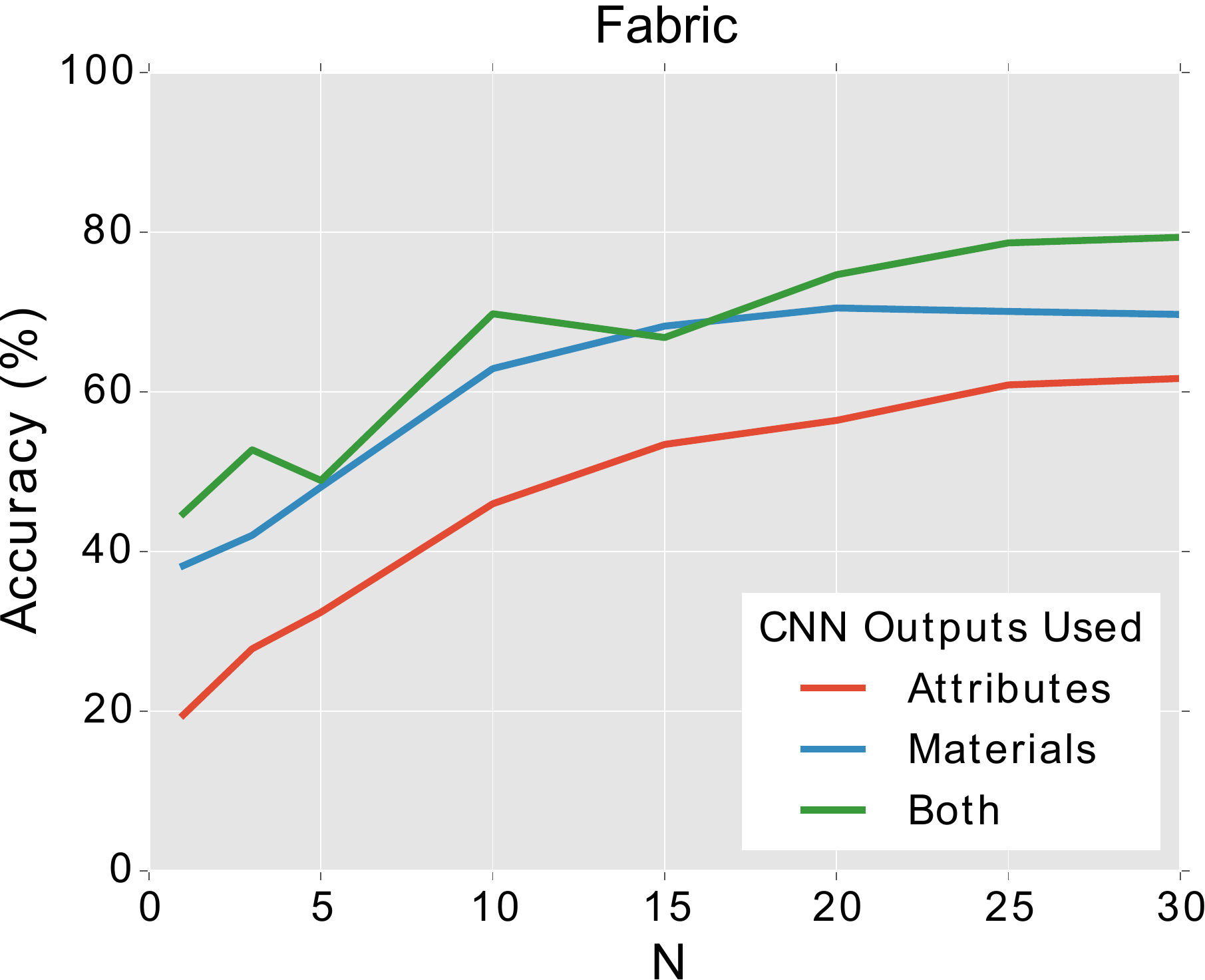}\hspace{12pt}\includegraphics[width=0.61\columnwidth]{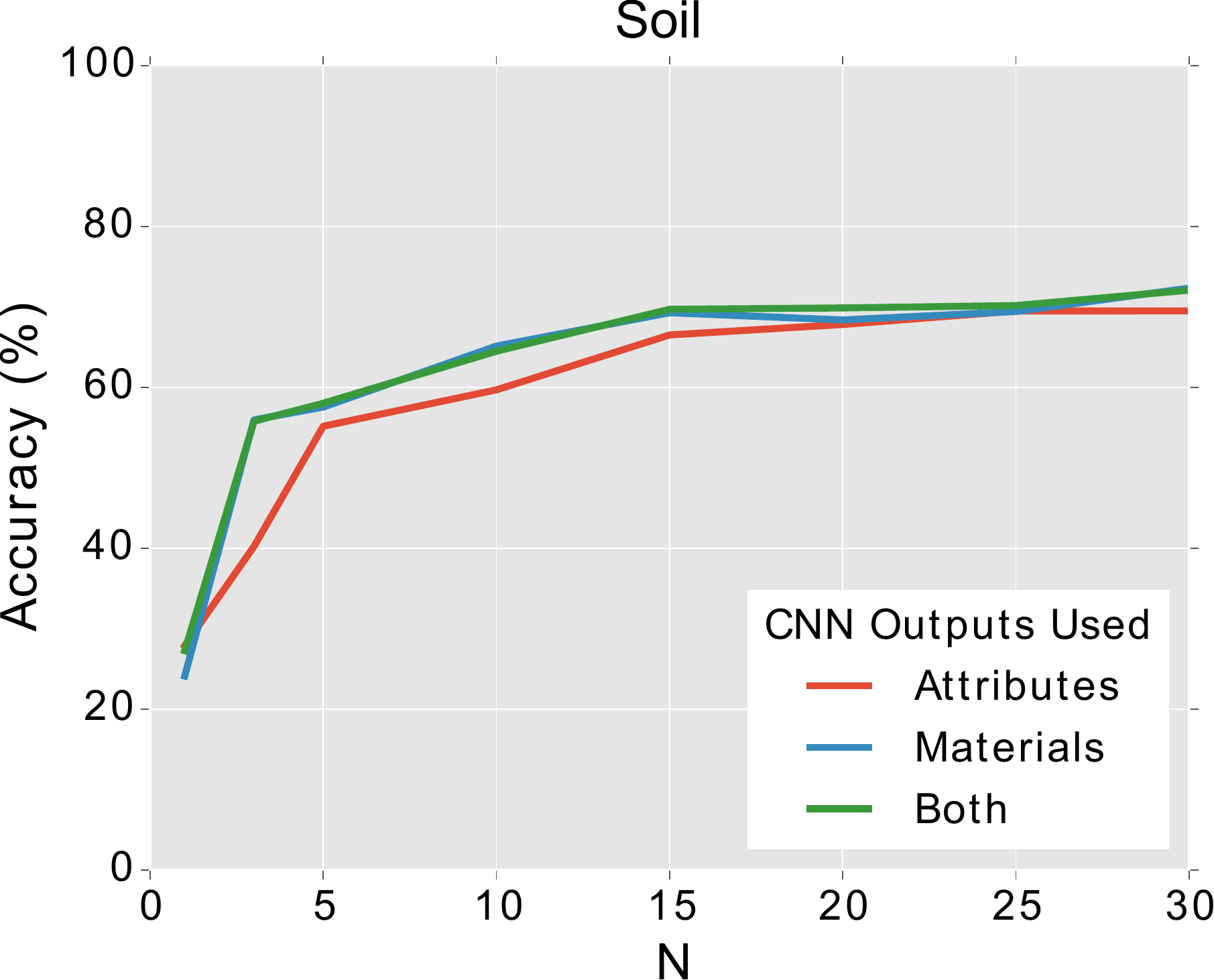}\hspace{12pt}\includegraphics[width=0.61\columnwidth]{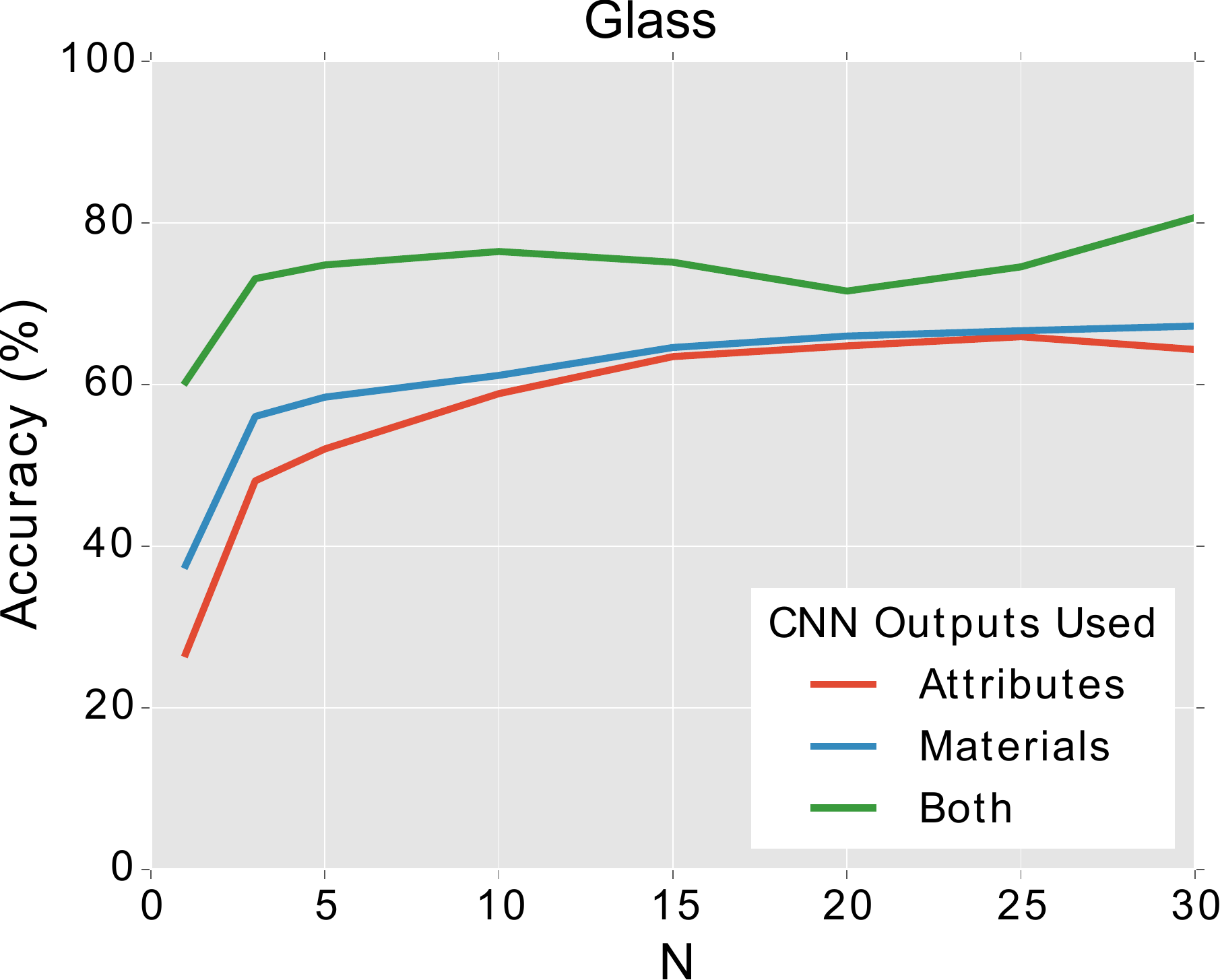} 
\par\end{centering}
\caption{\label{fig:Transfer-Learning-Accuracy}Graphs of novel category recognition
accuracy vs. training set size for various held-out categories. The
rapid plateau shows that we need only a small number of examples to
define a previously-unseen category. The accuracy difference between
feature sets shows that the attributes are contributing novel information.}
\end{figure*}

\figword\ref{fig:Transfer-Learning-Accuracy} shows plots of novel
category recognition effectiveness as the number of training examples
for the held-out category varies. We can see that the accuracy plateaus
quickly, indicating that the attributes provide a compact and accurate
representation for novel material categories. The number of images
we are required to extract patches from to obtain reasonable accuracy
is generally quite small (on the order of 10) compared to full material
category recognition frameworks which require hundreds of examples.
Furthermore, we include accuracy for the same predictions based on
only material probabilities instead of attribute probabilities, as
well as using a concatenation of both. This clearly shows that the
extracted  attributes can expose novel information in the MAC-CNN
that would not ordinarily be available.

\section{Conclusion}

Material properties provide valuable cues to guide our everyday interactions
with the materials that exhibit them. We aimed to recognize such properties
from images, in the form of visual material attributes, so that we
might make this information available for general scene understanding.
Our goal was not only to recognize these properties, but to do so
in a scalable fashion that allows our method to handle modern large-scale
material datasets.

By defining and recognizing a novel set of visual material attributes
\textendash{} material traits \textendash{} we showed that material
properties are locally-recognizable and can be aggregated to classify
materials in image regions. To scale the annotation process, we introduced
a novel method that probes our own perception of material appearance,
using only weak supervision in the form of yes/no similarity annotations,
to discover unnamed visual material attributes that serve the same
function as fully-supervised material traits. Our proposed MAC-CNN
allows us to apply our attribute discovery framework to modern large-scale
image databases in a seamless end-to-end fashion. Furthermore, the
design of the MAC-CNN exposes interesting parallels between human
and computer vision.

\section*{Acknowledgments}

This work was supported by the Office of Naval Research grants N00014-16-1-2158
(N00014-14-1-0316) and N00014-17-1-2406, and the National Science
Foundation awards IIS-1421094 and IIS-1715251. The Titan X used for
part of this research was donated by the NVIDIA Corporation.

\bibliographystyle{IEEEtran}
\bibliography{journal}
\vspace{-30pt}
\begin{IEEEbiography}[{\includegraphics[width=1\columnwidth]{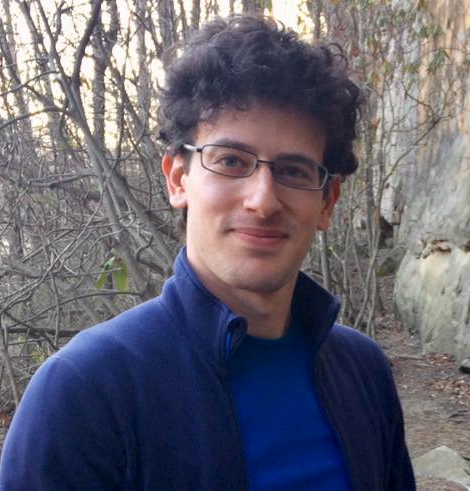}}]{Gabriel Schwartz}
is a Ph.D.~student at Drexel University in the Department of Computer
Science. He received his B.S.~and M.S~in Computer Science from Drexel
University in 2011. His research interests focus on computer vision
and machine learning.
\end{IEEEbiography}

\vspace{-30pt}
\begin{IEEEbiography}[{\includegraphics[width=1\columnwidth]{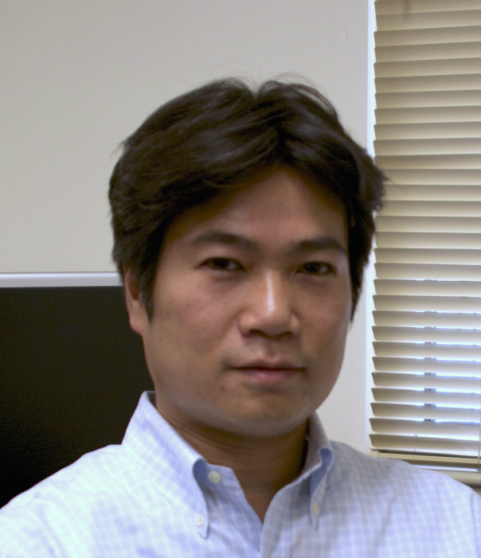}}]{Ko Nishino}
is a professor in the Department of Computing in the College of
Computing and Informatics at Drexel University. He is also an adjunct
associate professor in the Computer and Information Science Department
of the University of Pennsylvania and a visiting associate professor
of Osaka University. He received a B.E. and an M.E. in Information
and Communication Engineering in 1997 and 1999, respectively, and
a PhD in Computer Science in 2002, all from The University of Tokyo.
Before joining Drexel University in 2005, he was a Postdoctoral Research
Scientist in the Computer Science Department at Columbia University.
His primary research interests lie in computer vision and include
appearance modeling and synthesis, geometry processing, and video
analysis. His work on modeling eye reflections received considerable
media attention including articles in New York Times, Newsweek, and
NewScientist. He received the NSF CAREER award in 2008.
\end{IEEEbiography}

\end{document}